\documentclass{article}

\usepackage{microtype}
\usepackage{graphicx}
\usepackage{subfigure}
\usepackage{booktabs} 

\usepackage{hyperref}

\usepackage[accepted]{icml2025}

\usepackage{amsmath}
\usepackage{amssymb}
\usepackage{mathtools}
\usepackage{amsthm}
\usepackage{enumitem}

\usepackage[capitalize,noabbrev]{cleveref}

\theoremstyle{plain}
\newtheorem{theorem}{Theorem}[section]
\newtheorem{proposition}[theorem]{Proposition}

\theoremstyle{definition}

\theoremstyle{remark}

\usepackage[textsize=tiny]{todonotes}

\icmltitlerunning{BoA: Attention-aware Post-training Quantization without Backpropagation}

\usepackage{setspace}
\usepackage{multirow, makecell}
\usepackage{multicol}
\usepackage{algorithm}
\usepackage{threeparttable}

\usepackage{amsmath,amsfonts,bm}

\def\eqref#1{equation~\ref{#1}}

\def\1{\bm{1}}

\def\vw{{\bm{w}}}

\DeclareMathAlphabet{\mathsfit}{\encodingdefault}{\sfdefault}{m}{sl}
\SetMathAlphabet{\mathsfit}{bold}{\encodingdefault}{\sfdefault}{bx}{n}

\def\gO{{\mathcal{O}}}

\def\gQ{{\mathcal{Q}}}

\newcommand{\eg}{\emph{e.g.}}
\newcommand{\ie}{\emph{i.e.}}

\newcommand{\brecq}{\textsc{Brecq}}
\newcommand{\boa}{\textsc{BoA}}  

\newcommand{\mha}{\operatornamewithlimits{MHA}}
\newcommand{\vect}{\operatornamewithlimits{vec}}
\newcommand{\chol}{\operatornamewithlimits{Chol}}
\newcommand{\rope}{\operatornamewithlimits{RoPE}}

\usepackage{multirow}   
\usepackage{rotating}
\crefname{equation}{}{}

\begin{document}

\twocolumn[

\icmltitle{%
  \begin{tabular}{@{}c@{}l} 
    \raisebox{-0.05\totalheight}{\includegraphics[width=.33cm]{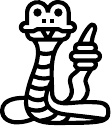}} &
    \begin{tabular}{@{}c@{}}~\boa: Attention-aware Post-training Quantization without Backpropagation 
    \end{tabular}
  \end{tabular}
}

\icmlsetsymbol{equal}{*}

\begin{icmlauthorlist}
\icmlauthor{Junhan Kim}{equal,yyy}
\icmlauthor{Ho-young Kim}{equal,yyy}
\icmlauthor{Eulrang Cho}{yyy}
\icmlauthor{Chungman Lee}{yyy}
\icmlauthor{Joonyoung Kim}{yyy}
\icmlauthor{Yongkweon Jeon}{yyy}
\end{icmlauthorlist}

\icmlaffiliation{yyy}{Samsung Research, Seoul, Republic of Korea}
\icmlcorrespondingauthor{Yongkweon Jeon}{dragwon.jeon@samsung.com}

\icmlkeywords{Machine Learning, ICML}

\vskip 0.3in
]



\printAffiliationsAndNotice{\icmlEqualContribution} 

\begin{abstract}

Post-training quantization (PTQ) is a promising solution for deploying large language models (LLMs) on resource-constrained devices. 
Early methods developed for small-scale networks, such as ResNet, rely on gradient-based optimization, which becomes impractical for hyper-scale LLMs with billions of parameters.
While recently proposed backpropagation-free or transformation-based methods alleviate this issue, they ignore inter-layer interactions or use the na\"ive nearest-rounding-based quantized weight assignment to save the heavy computational cost of weight optimization.
In this paper, we introduce a novel backpropagation-free PTQ algorithm that optimizes quantized weights by considering inter-layer dependencies. 
The key innovation is the development of attention-aware Hessian matrices that capture inter-layer interactions within the attention module. 
Extensive experiments demonstrate that our approach not only outperforms existing weight quantization methods but also shows good synergy with conventional methods to suppress activation outliers, leading to state-of-the-art weight-activation quantization performance.
The code will be available at \url{https://github.com/SamsungLabs/BoA}.

\end{abstract}

\section{Introduction}  \label{sec:intro}

The explosive growth in complexity (parameters) of large language models (LLMs)~\citep{touvron2023llama,zhang2022opt} has resulted in a proportional increase in computational costs, which has prompted an urgent need for efficient model processing and compression strategies.
Quantization has emerged as a pivotal solution and an essential step for deploying AI models on resource-constrained devices that primarily support fixed-point arithmetic. 

Two main categories of quantization approaches have been proposed to preserve the performance of original models: quantization-aware training (QAT)~\cite{jin2021kdlsq,liu2023llmqat} and post-training quantization (PTQ)~\cite{nagel2020up,kim2024towards}.
Although QAT can potentially outperform PTQ, its practicality diminishes when handling hyper-scale LLMs featuring billions of parameters.
Consequently, recent quantization efforts have been directed toward PTQ.

Although classic PTQ methods have successfully quantized small models such as ResNet~\citep{nagel2020up,li2021brecq,jeon2022mr}, they rely on time-consuming gradient-based optimization, so their efficacy decreases when the complexity of LLMs increases.
Accordingly, recent efforts have shifted to develop (a) backpropagation-free methods that optimize quantized weights based on Hessian~\cite{frantar2023optq} or (b) transformation-based methods that convert a model to be robust to quantization by applying smoothing, rotation, and permutation~\cite{shao2023omniquant, ashkboos2024quarot, liu2024spinquant, lin2024duquant}.
However, their weight quantization performance is limited as they ignore inter-layer dependencies or use na\"ive nearest rounding for weight quantization.

In this paper, we propose a novel backpropagation-free PTQ algorithm that considers inter-layer dependencies in optimizing quantized weights.
Our contributions are as follows:
\begin{itemize}
    \item We propose a novel PTQ algorithm called \boa\footnote{ \underline{B}ackpropagation-free \underline{o}ptimization for \underline{A}ttention-aware PTQ}.
    The key contribution is to exploit the attention reconstruction error, not the layer-wise reconstruction error, in approximating the Hessian to consider inter-layer dependencies within the attention module (\textbf{\cref{subsec:attention-considering-hessian}}).
    To our knowledge, \boa \ is the first method to optimize quantized weights by considering inter-layer dependencies without relying on gradient-based optimization.
    \item While the proposed Hessian facilitates the consideration of inter-layer dependencies, it requires additional memory and computations.
    To mitigate the overhead, we propose several techniques, including Hessian relaxation, efficient computation of inverse Hessians, and head-wise simultaneous quantization (\textbf{\cref{subsec:bptq}}).
    \item We evaluate \boa \ on publicly available LLMs. 
    From extensive experiments, we show that \boa \ outperforms existing backpropagation-free weight quantization methods by a significant margin, particularly for low-bit precision (\eg, INT2) (\textbf{\cref{subsec:weight-only results}}).
    Furthermore, when combined with existing methods to suppress outliers~\cite{ashkboos2024quarot, liu2024spinquant}, \boa \ achieves the state-of-the-art performance for both weight-only and weight-activation quantization (\textbf{Sections~\ref{subsec:weight-only results} and~\ref{subsec:weight-activation results}}).
\end{itemize}

\section{Related Works}  \label{sec:background}

When calibration data are available, PTQ aims to minimize the increase in task loss incurred by quantization.
Assuming the convergence of a network and the independence between layers, the problem of quantizing weights to minimize task loss degradation can be formulated as the following layer-wise reconstruction problem~\citep{nagel2020up}:
\begin{align}
    \min_{\Delta \mathbf{W}^{(\ell)}}~& \left \| \left ( \gQ (\mathbf{W}^{(\ell)}) - \mathbf{W}^{(\ell)} \right ) \mathbf{X}^{(\ell-1)} \right \|_{F}^{2}, \label{eq:layer-wise reconstruction}
\end{align}

where $\mathbf{X}^{(\ell-1)}$ is the input to the $\ell$-th layer parameterized by $\mathbf{W}^{(\ell)}$ and $\gQ$ is a quantization function. 
For a uniform quantization, if the nearest quantization bin is assigned to each weight, $\gQ$ is defined as
\begin{align*}
    \gQ(x)
        &= s \left ( \text{clamp} \left ( \left \lfloor \frac{x}{s} \right \rceil + z, 0, 2^{n}-1 \right ) - z \right ),
\end{align*}

where $s, z, n$ are the scale, zero-point, and bit-width, respectively, and $\lfloor \cdot \rceil$ represents the round-off.

Early works aimed to optimize the weight-rounding mechanism~\citep{nagel2020up, hubara2021accurate, li2021brecq,jeon2023genie}.
Instead of allocating the nearest integer, these studies attempted to assign integer weights that minimize the reconstruction error.
One popular approach is AdaRound, which learns quantized weights satisfying~\cref{eq:layer-wise reconstruction} via backpropagation~\citep{nagel2020up}.
This algorithm has been extended to \textsc{Brecq} where the block-wise reconstruction error has been used, instead of the layer-wise reconstruction error, to consider the inter-layer dependencies~\citep{li2021brecq}.
Although AdaRound and \brecq \ have successfully quantized small-sized models, they rely on time-consuming gradient-based optimizations, which renders their application to LLMs with billions of parameters challenging.
Consequently, recent efforts have shifted towards the development of cost-effective quantization methods for LLMs. 

These efforts can be classified into two orthogonal classes: \textbf{(a)} backpropagation-free Hessian-based integer weight optimization methods (\eg, GPTQ~\citep{frantar2023optq}) and \textbf{(b)} transformation-based methods that convert a model to be more robust to quantization by applying smoothing (\eg, SmoothQuant~\citep{xiao2023smoothquant}, OmniQuant~\citep{shao2023omniquant}, and AffineQuant~\citep{ma2024affinequant}), rotation (\eg, QuaRot~\citep{ashkboos2024quarot} and SpinQuant~\citep{liu2024spinquant}), or permutation (\eg, DuQuant~\citep{lin2024duquant}).

The proposed \boa \ belongs to the class (a) in that it optimizes quantized weights using the Hessian without relying on backpropagation.
Furthermore, similar to GPTQ, \boa \ can be combined with transformation-based methods such as QuaRot and SpinQuant (see \cref{sec:experiments}).
The primary difference over GPTQ lies in our optimization objective: while GPTQ assumes layer-wise independence and focuses on layer-wise reconstruction (which leads to performance degradation), our approach explicitly aims to preserve the attention output, enabling us to account for inter-layer dependencies within the attention module.

Other algorithms exploiting different strategies have also been proposed.
For example, SpQR~\citep{dettmers2023spqr}, SqueezeLLM~\citep{kim2023squeezellm}, and OAC~\citep{edalati2024oac} proposed a mixed-precision approach that assigns a large bit-width to quantization-sensitive weights or retains them in full-precision.
Compared to the standard uniform quantization, these algorithms require additional processing and memory costs in the inference and need special hardware and dedicated kernels without which accelerating the inference may not be easy.
Furthermore, unlike server-grade GPUs, on-device NPUs (\eg~Qualcomm Hexagon) lack support for the mixed precision format, and customizing kernels for desired functionalities is very challenging.
Thus, we exclude these algorithms in our comparison and focus on the more universally supported uniform quantization format.
We also exclude recent vector quantization approaches (\eg~QuIP\#~\cite{quip2} and AQLM~\cite{aqlm}) because they need additional memory (bits) for storing codebooks, which are required to perform the dequantization during the inference.

\section{Method}

\subsection{Overview of Proposed \boa}  \label{subsec:review of GPTQ}

The proposed \boa \ quantizes weights by repeating quantization and weight-update steps; once \boa \ quantizes one weight, it updates the remaining (not-yet-quantized) weights to compensate for the task loss degradation caused by the quantization.
The update formula to compensate for the quantization of the $q$-th weight $w_{q}$ is formulated as~\citep{frantar2023optq}
\begin{align} \label{eq:weight-update}
    \bm{\delta} \mathbf{w} = \frac{\gQ(w_{q}) - w_{q}}{[ \mathbf{U} ]_{q, q}} [ \mathbf{U} ]_{q, :} \text{ where } \mathbf{U}
        &= \chol ( \mathbf{H}^{-1} )^{T},
\end{align}
where $\mathbf{H}$ is the Hessian and $\chol(\cdot)$ denotes a Cholesky decomposition (\ie, $\mathbf{U}$ is an upper triangular matrix satisfying $\mathbf{H}^{-1} = \mathbf{U}^{T} \mathbf{U}$).

The key difference over GPTQ lies in the approximation of the Hessian $\mathbf{H}$.
In GPTQ, the layer-wise reconstruction error $\| \Delta \mathbf{W}^{(\ell)} \mathbf{X}^{(\ell-1)} \|_{F}^{2}$ in~\cref{eq:layer-wise reconstruction} has been used (\ie, the layer-wise independence has been assumed) to approximate $\mathbf{H}$ which yields the following Hessian equation\footnote{The second-order derivative of $\| \mathbf{M}_{1} \Delta \mathbf{W} \mathbf{M}_{2} \|_{F}^{2}$ with respect to $\Delta \mathbf{w}$ is $2 \mathbf{M}_{2} \mathbf{M}_{2}^{T} \otimes \mathbf{M}_{1}^{T} \mathbf{M}_{1}$ (see \cref{appendix:hessian_derivation} for the proof).  \label{footnote:hessian}}:
\begin{align} \label{eq:hessian_optq}
    \mathbf{H}^{(\mathbf{w}^{(\ell)})} 
        &\approx 2\mathbf{X}^{(\ell-1)} \mathbf{X}^{(\ell-1)^{T}} \otimes \mathbf{I},
\end{align}
where $\mathbf{w}^{(\ell)}$ is the flattened representation of $\mathbf{W}^{(\ell)}$, $\mathbf{H}^{(\mathbf{w}^{(\ell)})}$ is the Hessian for the $\ell$-th layer, $\otimes$ denotes the Kronecker product operation, and $\mathbf{I}$ is the identity matrix.
The approximated Hessian $\mathbf{H}^{(\mathbf{w}^{(\ell)})}$ in GPTQ relies solely on the input, which means that GPTQ cannot consider the influence of other layers. 
In other words, GPTQ neglects inter-layer dependencies within the attention module, a crucial aspect of Transformers, which results in limited performance~\citep{jeon2023frustratingly}. 
To overcome this, we develop Hessians that incorporate inter-layer dependencies and then use them instead of the conventional Hessian in~\cref{eq:hessian_optq}.

In \cref{tab:hessians}, we summarize the proposed Hessians; the detailed derivation is provided in the following subsections.
It can be observed that the proposed Hessians not only contain the term related to the input $\mathbf{X}$, but also involve the terms related to other layers (\eg, $\mathbf{K}_{h}^{T} \mathbf{K}_{h}$ for $\mathbf{W}_{Q}$).

\begin{table}[!t]
    \renewcommand{\arraystretch}{1.0}
    \footnotesize
    \centering
    \vspace{-0.25cm}
    \caption{Approximated Hessians in GPTQ and the proposed \boa}
    \begin{threeparttable}
    \begin{tabular}{c c c}
        \toprule
        Method & Layer & $\mathbf{H} = \mathbf{H}_{\text{col}} \otimes \mathbf{H}_{\text{row}}$ \\
        \toprule
        GPTQ
        & $\mathbf{W}_{\{ Q, K, V \}}$ & $2 \mathbf{X} \mathbf{X}^{T} \otimes \mathbf{I}$ \\
        \midrule
        \multirowcell{4}{\bf{\boa}}
        & $\mathbf{W}_{Q, h}$ & $2 \mathbf{X} \mathbf{X}^{T} \otimes \mathbf{K}_{h}^{T} \mathbf{K}_{h}$ \\
        & $\mathbf{W}_{K, h}$ & $2 \mathbf{X} \mathbf{X}^{T} \otimes \mathbf{Q}_{h}^{T} \mathbf{Q}_{h}$ \\
        & $\mathbf{W}_{V, h}$ & $2 \mathbf{X} \mathbf{A}_{h}^{T} \mathbf{A}_{h} \mathbf{X}^{T} \otimes \mathbf{W}_{\text{out}, h}^{T} \mathbf{W}_{\text{out}, h}$ \\
        & $\mathbf{W}_{\text{out}, h}$ & $2\mathbf{X}_{\text{out}, h} \mathbf{X}_{\text{out}, h}^{T} \otimes \mathbf{I}$ \\
        \bottomrule
    \end{tabular}
    \begin{tablenotes}
        \item[*] For $\mathbf{W}_{Q, h}$ and $\mathbf{W}_{K, h}$ of models exploiting rotary position embedding, see~\cref{eq:relaxed-hessian_query_rope} and~\cref{eq:relaxed-hessian_key_rope}.
    \end{tablenotes}
    \end{threeparttable}
    \label{tab:hessians}
    \vspace{-0.25cm}
\end{table}

\begin{algorithm*}[t]
\begin{spacing}{1.05}
\caption{\boa} 
\footnotesize
\label{algo:boa}
\renewcommand\algorithmicrequire{\textbf{Input}:}
\renewcommand\algorithmicensure{\textbf{Output}:}
\begin{algorithmic}[1]
\REQUIRE weights $\mathbf{W}_{\{ Q, K, V \}} \in \mathbb{R}^{H \times d_{h} \times d}$ and inputs $\mathbf{X}$ of the Transformer layer
    \FOR{$\mathbf{W} \in \{ \mathbf{W}_{Q}, \mathbf{W}_{K}, \mathbf{W}_{V} \} $}
    \STATE Initialize quantized output: $\mathbf{Q} \leftarrow \mathbf{0}_{H \times d_{h} \times d}$
    \STATE Initialize (row-wise) quantization errors: $\mathbf{E} \leftarrow \mathbf{0}_{H \times d}$
    \STATE Compute attention-aware Hessians: $\mathbf{H}_{h} = \mathbf{H}_{\text{col}, h} \otimes \mathbf{H}_{\text{row}, h}$~~~~~~~~~~~~~~~~~~~~~~~~~~~$\triangleright$ see \cref{tab:hessians}
    \STATE Set step size (scale) $\mathbf{S}$: $\min_{\mathbf{S}} \hspace{.2mm} \text{tr} \hspace{-.5mm} \left ( \Delta \mathbf{W}_{h} \mathbf{H}_{\text{col}, h} \Delta \mathbf{W}_{h}^{T} \right )$
    \STATE Compute inverse Hessians $\mathbf{H}_{\text{col}, h}^{-1}$ and $\mathbf{H}_{\text{row}, h}^{-1}$
    \STATE Compute $\mathbf{U}_{\text{col}, h} = \chol(\mathbf{H}_{\text{col}, h}^{-1})^{T}$ and $\mathbf{U}_{\text{row}, h} = \chol(\mathbf{H}_{\text{row}, h}^{-1})^{T}$
    \FOR{$j=0, \ldots, d_{h} - 1$}
    \STATE Construct $\mathbf{W}^{(j)} \in \mathbb{R}^{H \times d}$ by stacking the $j$-th rows $[ \mathbf{W}_{h} ]_{j, :}$
    \STATE Quantize $\mathbf{W}^{(j)}$:  $(\mathbf{Q}_{:, j, :}, \mathbf{E}) \leftarrow \text{GPTQ}(\mathbf{W}^{(j)}, \mathbf{U}_{\text{col}, h}, \mathbf{S})$~~~~~~~~~~~~~~~~~~~~~~~~~~~~$\triangleright$ see \cref{appendix:pseudocode-gptq}
    \STATE Update remaining rows:
    $[\mathbf{W}_{h}]_{j:,:} \leftarrow [\mathbf{W}_{h}]_{j:,:} - \frac{[\mathbf{U}_{\text{row}, h}^{T}]_{j:, j} \cdot \mathbf{E}_{h, :} \cdot \mathbf{U}_{\text{col}, h}}{[\mathbf{U}_{\text{row}, h}]_{j, j}}$ ~~~\hspace{.4mm}$\triangleright$ see \cref{prop:refined_update} 
    \ENDFOR
    \ENDFOR
\ENSURE quantized weights $\mathbf{Q}$
\end{algorithmic}
\end{spacing}
\end{algorithm*}

\subsection{Proposed Attention-aware Hessian}  \label{subsec:attention-considering-hessian}

To consider the inter-layer dependencies within the attention module, we exploit the attention reconstruction error rather than the layer-wise reconstruction error when approximating the Hessian.
For an input sequence $\mathbf{X} \in \mathbb{R}^{d \times L}$, the output of the multi-head attention (MHA) is expressed as
\begin{align} \label{eq:mha}
    \mha(\mathbf{X}) 
        \hspace{-1mm}= \hspace{-1mm} \sum_{h=1}^{H} \hspace{-.5mm} \mathbf{W}_{\text{out}, h} (\mathbf{A}_{h} \mathbf{V}_{h})^{T} \hspace{-.7mm}, 
    \mathbf{A}_{h} 
        \hspace{-1mm}= \hspace{-.7mm} \sigma \hspace{-.7mm} \left ( \frac{\mathbf{Q}_{h} \mathbf{K}_{h}^{T}}{\sqrt{d_{h}}} \right ) \hspace{-.7mm}, 
\end{align}
where $\mathbf{Q}_{h}, \mathbf{K}_{h}, \mathbf{V}_{h} \in \mathbb{R}^{L \times d_{h}}$ are query, key, value for the $h$-th attention head, $d_{h}$ is the head dimension, $\sigma$ is the row-wise softmax function, and $H$ is the total number of heads.

\paragraph{Hessian for query}
When quantizing the query projection weights $\mathbf{W}_{Q, h}$, $\mathbf{W}_{\text{out}, h}$ and $\mathbf{V}_{h}$ remain unchanged, but the attention weights $\mathbf{A}_{h}$ change.
Using the first-order Taylor polynomial, the perturbation in $\mathbf{A}_{h}$ can be approximated as
\begin{align}
    \Delta \mathbf{A}_{h}
        &= \sigma \left ( \frac{ (\mathbf{Q}_{h} + \Delta \mathbf{Q}_{h}) \mathbf{K}_{h}^{T}}{\sqrt{d_{h}}} \right ) - \sigma \left ( \frac{ \mathbf{Q}_{h} \mathbf{K}_{h}^{T}}{\sqrt{d_{h}}} \right ) \nonumber \\
        &\approx \frac{\Delta \mathbf{Q}_{h} \mathbf{K}_{h}^{T}}{\sqrt{d_{h}}} \mathbf{J}_{\sigma}^{T}
        = \frac{\mathbf{X}^{T} \Delta \mathbf{W}_{Q, h}^{T} \mathbf{K}_{h}^{T} \mathbf{J}_{\sigma}^{T}}{\sqrt{d_{h}}},  \label{eq:approximation_attn-weights_query}
\end{align}
where $\mathbf{J}_{\sigma}$ is the Jacobian matrix of the softmax function $\sigma$.
Thus, the attention reconstruction error is expressed as
\begin{align}
    \| \hspace{-.2mm} \Delta \hspace{-.5mm} \mha \hspace{-.2mm} ( \mathbf{X} ) \hspace{-.2mm} \|_{F}^{2}
        &= \| \mathbf{W}_{\text{out}, h} (\Delta \mathbf{A}_{h} \mathbf{V}_{h})^{T} \|_{F}^{2} \nonumber \\
        &\approx \left \| \frac{\mathbf{W}_{\text{out}, h} \mathbf{V}_{h}^{T} \mathbf{J}_{\sigma} \mathbf{K}_{h}}{\sqrt{d_{h}}} \Delta \mathbf{W}_{Q, h} \mathbf{X} \right \|_{F}^{2}, \label{eq:attn-recon-error_query}
\end{align}
which yields the following Hessian for $\mathbf{W}_{Q, h}$ (see \cref{footnote:hessian}):
\begin{align}  \label{eq:hessian_query}
    \mathbf{H}^{(\mathbf{w}_{Q, h}\hspace{-.1mm})}
        &\hspace{-1mm}= \hspace{-.7mm}
        2 \mathbf{X} \mathbf{X}^{\hspace{-.3mm}T} 
        \hspace{-1mm} \otimes \hspace{-1mm}
        \frac{\mathbf{K}_{h}^{T} \mathbf{J}_{\sigma}^{T} \mathbf{V}_{\hspace{-.6mm} h} \hspace{-.6mm} \mathbf{W}_{\text{out}, h}^{T} \hspace{-.6mm} \mathbf{W}_{\text{out}, h} \hspace{-.6mm}\mathbf{V}_{\hspace{-.6mm} h}^{\hspace{-.1mm}T} \mathbf{J}_{\sigma} \mathbf{K}_{h}}{d_{h}}\hspace{-.5mm}.
\end{align}

\paragraph{Hessian for key}
As in the quantization of the query projection weights, the attention weight $\mathbf{A}_{h}$ changes when quantizing the key projection weights $\mathbf{W}_{K, h}$.
By following the steps for~\cref{eq:approximation_attn-weights_query}, $\mathbf{A}_{h}$ can be approximated as
\begin{align*}
    \Delta \mathbf{A}_{h}
        &\hspace{-.5mm}\approx \frac{\mathbf{Q}_{h} \Delta \mathbf{K}_{h}^{T}}{\sqrt{d_{h}}} \mathbf{J}_{\sigma}^{T}
        = \frac{\mathbf{Q}_{h} \Delta \mathbf{W}_{K, h} \mathbf{X}\mathbf{J}_{\sigma}^{T}}{\sqrt{d_{h}}},
\end{align*}
and then the attention reconstruction error is expressed as
\begin{align*}
    \| \hspace{-.2mm} \Delta \hspace{-.5mm} \mha \hspace{-.2mm} ( \mathbf{X} ) \hspace{-.2mm} \|_{F}^{2}
        \approx \left \| \hspace{-.2mm} \frac{\mathbf{Q}_{h}}{\sqrt{d_{h}}} \Delta \hspace{-.5mm}\mathbf{W}_{K, h} \mathbf{X} \mathbf{J}_{\sigma}^{T} \mathbf{V}_{h} \mathbf{W}_{\text{out}, h}^{T} \right \|_{F}^{2}.
\end{align*}
Thus, we obtain the following Hessian for $\mathbf{W}_{K, h}$:
\begin{align}  \label{eq:hessian_key}
    \mathbf{H}^{(\mathbf{w}_{K, h}\hspace{-.1mm})}
        &\hspace{-1mm}= \hspace{-.7mm} 
        2 \mathbf{X} \mathbf{J}_{\sigma}^{T} \mathbf{V}_{\hspace{-.6mm} h} \hspace{-.6mm} \mathbf{W}_{\text{out}, h}^{T} \hspace{-.6mm} \mathbf{W}_{\text{out}, h} \hspace{-.6mm}\mathbf{V}_{\hspace{-.6mm} h}^{\hspace{-.1mm}T} \mathbf{J}_{\sigma} \mathbf{X}^{\hspace{-.3mm}T} \hspace{-1mm} \otimes \hspace{-1mm} \frac{\mathbf{Q}_{h}^{T} \mathbf{Q}_{h}}{d_{h}}.
\end{align}

\paragraph{Hessian for value}
When quantizing the weights $\mathbf{W}_{V, h}$ of the value projection, only $\mathbf{V}_{h}$ changes.
Thus, we have
\begin{align*}
    \left \| \Delta \mha (\mathbf{X}) \right \|_{F}^{2}
        &= \| \mathbf{W}_{\text{out}, h} (\mathbf{A}_{h} \Delta \mathbf{V}_{h})^{T} \|_{F}^{2} \nonumber \\
        &= \| \mathbf{W}_{\text{out}, h} \Delta \mathbf{W}_{V, h} \mathbf{X} \mathbf{A}_{h}^{T} \|_{F}^{2},
\end{align*}
which yields the following Hessian for $\mathbf{W}_{V, h}$:
\begin{align}  \label{eq:hessian_value}
    \mathbf{H}^{(\mathbf{w}_{V, h})}
        &= 2 \mathbf{X} \mathbf{A}_{h}^{T} \mathbf{A}_{h} \mathbf{X}^{T} \otimes \mathbf{W}_{\text{out}, h}^{T} \mathbf{W}_{\text{out}, h}.
\end{align}

\paragraph{Hessian for out}

When the out-projection weights $\mathbf{W}_{\text{out}, h}$ are quantized, the attention reconstruction error is
\begin{align*}
    \left \| \Delta \mha (\mathbf{X}) \right \|_{F}^{2}
        &= \| \Delta \mathbf{W}_{\text{out}, h} ( \mathbf{A}_{h} \mathbf{V}_{h} )^{T} \|_{F}^{2}.
\end{align*}
Thus, the corresponding Hessian is
\begin{align}  \label{eq:hessian_out}
    \mathbf{H}^{(\mathbf{w}_{out, h})}
        &\hspace{-1mm}= \hspace{-.7mm} 2 \mathbf{V}_{h}^{T} \mathbf{A}_{h}^{T} \mathbf{A}_{h} \mathbf{V}_{h} \otimes \mathbf{I}
        \hspace{-1mm}= \hspace{-.7mm} 2 \mathbf{X}_{\text{out}, h} \mathbf{X}_{\text{out}, h}^{T} \otimes \mathbf{I},
\end{align}
where $\mathbf{X}_{\text{out}, h} = (\mathbf{A}_{h} \mathbf{V}_{h})^{T}$.

\subsection{Efficient Implementation of \boa}  \label{subsec:bptq}

While inter-layer dependencies within the attention module can be considered by exploiting the proposed Hessians, they are significantly more complex than the conventional Hessian in~\cref{eq:hessian_optq}, which may incur high computational costs.
For example, computing the proposed Hessians in~\cref{eq:hessian_query} and~\cref{eq:hessian_key} would be more expensive than computing the conventional one in~\cref{eq:hessian_optq}.
In this subsection, we present techniques to mitigate the computational overheads incurred by the proposed attention-aware Hessians.

\paragraph{Hessian relaxation}
The largest overhead related to the computation of the proposed Hessians is the Jacobian matrix $\mathbf{J}_{\sigma}$ in~\cref{eq:hessian_query} and~\cref{eq:hessian_key}.
For an input sequence of length $L$, the shape of $\mathbf{J}_{\sigma}$ is $H \times L \times L \times L$, which requires a large amount of memory and high computational cost (\eg, more than 400~GB even for the OPT-125M model when $L=2048$). 

To mitigate such overhead, we establish a relaxed Hessian that does not require computing $\mathbf{J}_{\sigma}$.
To this end, we build the following upper bound for the attention reconstruction error in~\cref{eq:attn-recon-error_query}, which will be used as its surrogate:
\begin{align*}
    \| \Delta \mha ( \mathbf{X} ) \|_{F}^{2}
        &\hspace{-.5mm}\le\hspace{-.5mm} \left \| \frac{\mathbf{W}_{\text{out}, h} \mathbf{V}_{h}^{T} \mathbf{J}_{\sigma}}{\sqrt{d_{h}}}  \right \|_{F}^{2} \hspace{-1.5mm} \hspace{-.5mm} \cdot \hspace{-.3mm} \left \| \mathbf{K}_{h} \Delta \mathbf{W}_{Q, h} \mathbf{X} \right \|_{F}^{2}.
\end{align*}
Noting that the constant term $\| \mathbf{W}_{\text{out}, h} \mathbf{V}_{h}^{T} \mathbf{J}_{\sigma} \|_{F}^{2}$ does not affect quantization,\footnote{The update $\bm{\delta} \mathbf{w}$ in~\cref{eq:weight-update} is not affected by the constant multiple of $\mathbf{H}$ because $[ c\mathbf{U} ]_{q, :} / [ c\mathbf{U} ]_{q, q} \hspace{-.5mm} = \hspace{-.5mm} [ \mathbf{U} ]_{q, :} / [ \mathbf{U} ]_{q, q}$ for any constant $c$.} we use the term $\| \mathbf{K}_{h} \Delta \mathbf{W}_{Q, h} \mathbf{X} \|_{F}^{2}$ as a surrogate of the attention reconstruction error when deriving the Hessian for $\mathbf{W}_{Q, h}$, which yields the following Hessian:
\begin{align}
    \mathbf{H}^{(\mathbf{w}_{Q, h})}
        &= 2 \mathbf{X} \mathbf{X}^{T} \otimes \mathbf{K}_{h}^{T} \mathbf{K}_{h}.  \label{eq:relaxed-hessian_query}
\end{align}
Similarly, we can establish a relaxed Hessian for $\mathbf{W}_{K, h}$ as
\begin{align}
    \mathbf{H}^{(\mathbf{w}_{K, h})}
        &= 2 \mathbf{X} \mathbf{X}^{T} \otimes \mathbf{Q}_{h}^{T} \mathbf{Q}_{h}.  \label{eq:relaxed-hessian_key}
\end{align}

We note that if rotary position embedding (RoPE)~\cite{rope} is used, the MHA output is different from that in~\cref{eq:mha}, and thus the corresponding attention-aware Hessians would also be different.
Specifically, since the attention weight $\mathbf{A}_{h}$ for models exploiting RoPE is expressed as
\begin{align} \label{eq:attention weight_rope}
    \mathbf{A}_{h} 
        \hspace{-1mm}= \hspace{-.7mm} \sigma \hspace{-.7mm} \left ( \frac{\widetilde{\mathbf{Q}}_{h} \widetilde{\mathbf{K}}_{h}^{T}}{\sqrt{d_{h}}} \right ) \hspace{-.7mm}, 
    \widetilde{\mathbf{Q}}_{h} 
        \hspace{-1mm}= \hspace{-.7mm} \rope(\mathbf{Q}_{h}), 
    \widetilde{\mathbf{K}}_{h} 
        \hspace{-1mm}= \hspace{-.7mm} \rope(\mathbf{K}_{h}),
\end{align}
the surrogate $\| \mathbf{K}_{h} \Delta \mathbf{Q}_{h}^{T} \|_{F}^{2}$ used to develop the attention-aware Hessian in~\cref{eq:relaxed-hessian_query} changes to $\| \widetilde{\mathbf{K}}_{h} \Delta \widetilde{\mathbf{Q}}_{h}^{T} \|_{F}^{2}$.
As a result, we obtain different Hessians for models exploiting RoPE (see \cref{appendix:hessian_derivation_rope} for the detailed derivation):
\begin{align}
    \mathbf{H}^{(\mathbf{w}_{Q, h})}
        &= 2 \mathbf{X} \mathbf{X}^{T} \otimes \frac{1}{L}\sum_{\ell=1}^{L} \mathbf{R}_{\ell}^{T} \widetilde{\mathbf{K}}_{h}^{T} \widetilde{\mathbf{K}}_{h} \mathbf{R}_{\ell},  \label{eq:relaxed-hessian_query_rope} \\
    \mathbf{H}^{(\mathbf{w}_{K, h})}
        &= 2 \mathbf{X} \mathbf{X}^{T} \otimes \frac{1}{L}\sum_{\ell=1}^{L} \mathbf{R}_{\ell}^{T} \widetilde{\mathbf{Q}}_{h}^{T} \widetilde{\mathbf{Q}}_{h} \mathbf{R}_{\ell},  \label{eq:relaxed-hessian_key_rope}
\end{align}
where $\mathbf{R}_{\ell}$ be the rotary matrix for the $\ell$-th token (see eq.~(15) in \cite{rope}).

We summarize the relaxed Hessians in \cref{tab:hessians}.

\paragraph{Efficient computation of inverse Hessians} 

After computing Hessians, their inverse matrices need to be computed to update the remaining weights after each quantization (see~\cref{eq:weight-update}). 
Owing to the size of the proposed Hessians being $dd_{h} \times dd_{h}$, the complexity of the computation of the inverse Hessian would be $\gO(d^{3}d_{h}^{3})$ in our approach.
This is considerably more expensive than the complexity $\gO(d^{3})$ of GPTQ, where the inverse of only $\mathbf{X} \mathbf{X}^{T} \in \mathbb{R}^{d \times d}$ in \cref{eq:hessian_optq} is needed~\citep{frantar2023optq}.

For the efficient inverse computation, we exploit the useful properties of the Kronecker product (see \cref{eq:Kron-prop_1}-\cref{eq:Kron-prop_3} in \cref{appendix:hessian_derivation}).
Specifically, let $\mathbf{H} = \mathbf{H}_{\text{col}} \otimes \mathbf{H}_{\text{row}}$ where $\mathbf{H}_{\text{col}} \in \mathbb{R}^{d \times d}$ and $\mathbf{H}_{\text{row}} \in \mathbb{R}^{d_{h} \times d_{h}}$, then we obtain
\begin{align*}
    \mathbf{H}^{-1}
        &= (\mathbf{H}_{\text{col}} \otimes \mathbf{H}_{\text{row}})^{-1}
        = \mathbf{H}_{\text{col}}^{-1} \otimes \mathbf{H}_{\text{row}}^{-1}.
\end{align*}
This implies that the inverse Hessian $\mathbf{H}^{-1}$ can be obtained by computing $\mathbf{H}_{\text{col}}^{-1}$ and $\mathbf{H}_{\text{row}}^{-1}$ (line~5 in \cref{algo:boa}) whose complexity is $\gO(d^{3}) + \gO(d_{h}^{3})$ ($=\gO(d^{3})$), not $\gO(d^{3}d_{h}^{3})$.
Similarly, we can efficiently perform the Cholesky decomposition (\ie, $\chol(\mathbf{H}^{-1})$ in \cref{eq:weight-update}) with the same order of complexity as in GPTQ.
Specifically, if $\mathbf{L}_{1} = \chol(\mathbf{H}_{\text{col}}^{-1})$ and $\mathbf{L}_{2} = \chol(\mathbf{H}_{\text{row}}^{-1})$, $\mathbf{H}^{-1}$ can be expressed as
\begin{align*}
    \mathbf{H}^{-1}
        &\hspace{-.5mm}=\hspace{-.5mm} \mathbf{L}_{1} \mathbf{L}_{1}^{T} \otimes \mathbf{L}_{2} \mathbf{L}_{2}^{T}
        \hspace{-.5mm}=\hspace{-.5mm} (\mathbf{L}_{1} \otimes \mathbf{L}_{2}) (\mathbf{L}_{1} \otimes \mathbf{L}_{2})^{T}.
\end{align*}
Subsequently, noting that the Kronecker product of lower triangular matrices is also lower triangular, we obtain
\begin{align*}
    \chol(\mathbf{H}^{-1})
        &= \mathbf{L}_{1} \otimes \mathbf{L}_{2}
        = \chol(\mathbf{H}_{\text{col}}^{-1}) \otimes \chol(\mathbf{H}_{\text{row}}^{-1}).
\end{align*}
Thus, we can obtain $\chol (\mathbf{H}^{-1})$ by computing $\chol(\mathbf{H}_{\text{col}}^{-1})$ and $\chol(\mathbf{H}_{\text{row}}^{-1})$ (line 6 in \cref{algo:boa}).
Consequently, the computational complexity of the Choleksy decomposition would be $\gO(d^{3})$, not $\gO(d^{3}d_{h}^{3})$.

\begin{figure}[t]
\centering
\includegraphics[width=.99\linewidth]{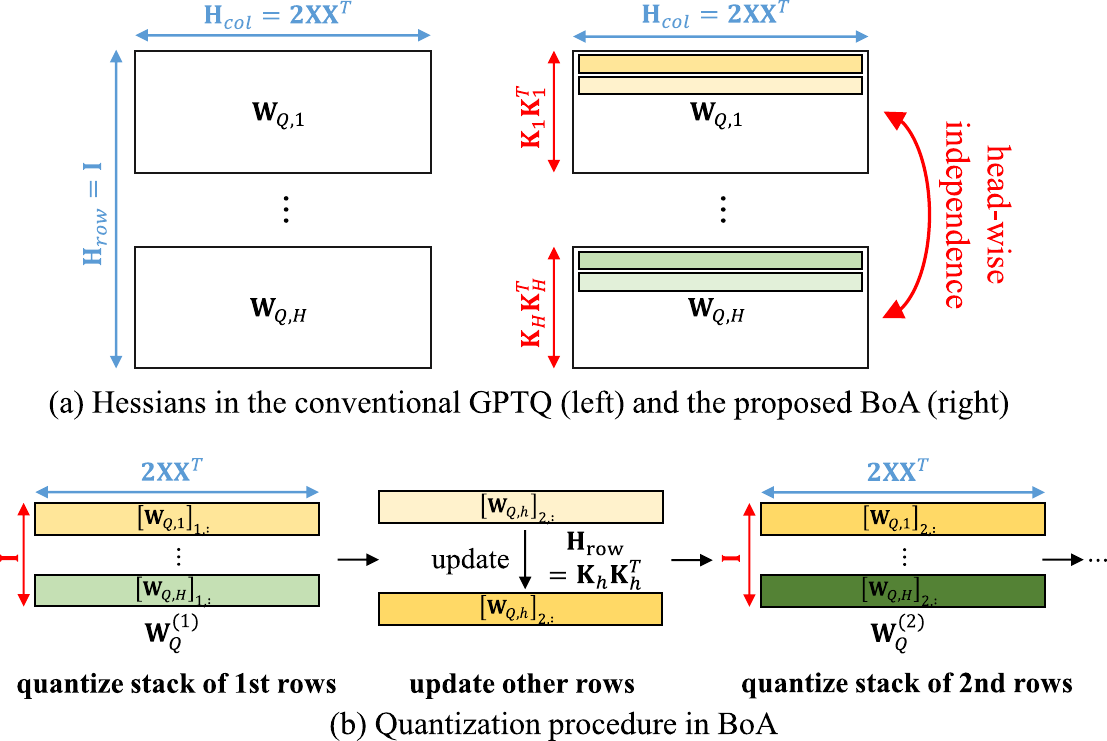}
\vspace{-0.2cm}
\caption{Illustration of \boa \ for the query projection $\mathbf{W}_{Q}$.}
\vspace{-0.2cm}
\label{fig:boa}
\end{figure}

\paragraph{Simultaneous quantization of different heads}
Because the proposed Hessians model the dependency between different rows (\eg, $\mathbf{H}_{\text{row}} = \mathbf{K}_{h}^{T} \mathbf{K}_{h}$ for $\mathbf{W}_{Q, h}$; see~\cref{eq:relaxed-hessian_query}), we can compensate for the quantization error of a certain row by updating other rows.
Specifically, the row-update formula is formulated as in the following proposition for given Hessian $\mathbf{H} = \mathbf{H}_{\text{col}} \otimes \mathbf{H}_{\text{row}}$.

\begin{proposition} \label{prop:refined_update}
    Let $\mathbf{W}_{h}$ be a $d_{h} \times d$ matrix whose Hessian is $\mathbf{H}_{h} = \mathbf{H}_{\text{col}, h} \otimes \mathbf{H}_{\text{row}, h}$. 
    Suppose $\mathbf{W}_{h}$ is quantized via \cref{eq:weight-update}.
    If the $j$-th row of $\mathbf{W}_{h}$ has been quantized, then the update formula to compensate for the quantization is expressed as
    \begin{align} \label{eq:refined_update}
        [\delta\mathbf{W}_{h}]_{j:, :}
        = - \frac{[\mathbf{U}_{\text{row}, h}^{T}]_{j:, j} \cdot \mathbf{e} \cdot \mathbf{U}_{\text{col}, h}}{[\mathbf{U}_{\text{row}, h}]_{j, j}},
    \end{align}
    where $\mathbf{U}_{\text{col}, h} = \chol (\mathbf{H}_{\text{col}, h}^{-1})^{T}$, $\mathbf{U}_{\text{row}, h} = \chol (\mathbf{H}_{\text{row}, h}^{-1})^{T}$, and $\mathbf{e} \in \mathbb{R}^{1 \times d}$ is the quantization error of the $j$-th row defined as $e_{i} = ([\mathbf{W}_{h}]_{j, i} - \mathcal{Q}([\mathbf{W}_{h}]_{j, i})) / [\mathbf{U}_{\text{col}, h}]_{i, i}$.
\end{proposition}
\begin{proof}
    See \cref{appendix:refined_weight_update}.
\end{proof}

The update formula in~\cref{eq:refined_update} implies that we do not need to compute and store the Cholesky decomposition $\mathbf{U}_{h} = \mathbf{U}_{\text{col}, h} \otimes \mathbf{U}_{\text{row}, h}$ of the full Hessian $\mathbf{H}_{h}$ for updating weights; only $\mathbf{U}_{\text{col}, h}$ and $\mathbf{U}_{\text{row}, h}$ are sufficient for updating weights, and thus we can save the memory cost caused by the high-dimensional Kronecker product operation.
\cref{prop:refined_update} also implies that the conventional GPTQ cannot compensate quantization error of certain row by updating other rows because $\mathbf{H}_{\text{row}} = \mathbf{I}$ (see~\cref{eq:hessian_optq}) and thus $\mathbf{U}_{\text{row}} = \mathbf{I}$ and $[\delta \mathbf{W}]_{i, :} = \mathbf{0}$ for all $i \neq j$.

While the proposed \boa \ can compensate for the quantization error of each row, the rows must be quantized sequentially (not simultaneously).
For example, the second row can be quantized after being updated to compensate for the quantization error of the first row.
To accelerate the quantization process, we assume independence between different attention heads (see \cref{fig:boa}(a)), under which rows related to different heads are independent and can thus be quantized together.
For a better understanding, we consider the query projection $\mathbf{W}_{Q}$ as an example (see \cref{fig:boa}(b)).
In the quantization step, we stack the $j$-th rows $[ \mathbf{W}_{Q, h} ]_{j, :}$ of all different heads, constructing the sub-weight matrix $\mathbf{W}_{Q}^{(j)} \in \mathbb{R}^{H \times d}$ (line~8 in \cref{algo:boa}).
Because the rows of $\mathbf{W}_{Q}^{(j)}$ are mutually independent, all the rows of $\mathbf{W}_{Q}^{(j)}$ can be quantized simultaneously as in GPTQ (line~9 in \cref{algo:boa}).
Following the quantization of $j$-th rows, we compensate for the quantization error by updating the remaining rows.
In this update step, we use the refined weight-update formula in \cref{eq:refined_update} (line~10 in \cref{algo:boa}).

\begin{table}[!t]
    \renewcommand{\arraystretch}{1.2}
    \scriptsize
    \centering
    \vspace{-.25cm}
    \caption{Processing time (hour) of \boa \ with and without simultaneous quantization of different heads}
    \begin{threeparttable}
    \begin{tabular}{c c c c}
    \toprule
    \multirowcell{2}{Simultaneous \\ Quantization} & \multicolumn{3}{c}{LLaMA Model Size} \\
    \cline{2-4}
    & 7B & 13B & 30B \\
    \toprule
    X & 27.75 & 51.66 & 135.4 \\
    \midrule
    O & \bf{0.961} & \bf{1.553} & \bf{3.295} \\
    \bottomrule
    \end{tabular}
    \end{threeparttable}
    \label{tab:processing_time_ablation}
    \vspace{-.25cm}
\end{table}

\begin{table*}[t]
    \renewcommand{\arraystretch}{1.0}
    \scriptsize
    \centering
    \caption{Weight-only quantization performance on LLaMA2 and LLaMA3 models without transformation}
    \begin{threeparttable}
    \begin{tabular}{c c c c c c c c c c c c c}
    \toprule
    \multirowcell{2}{Model} & \multirowcell{2}{Precision} & \multirowcell{2}{Method} & \multirowcell{2}{Wiki2 PPL ($\downarrow$)} & \multicolumn{9}{c}{Zero-shot Accuracy ($\uparrow$)} \\
    \cline{5-13}
    & & & & Arc-c & Arc-e & BQ & HS & LAMB & OBQA & PIQA & WG & Average \\
    \toprule
    \multirowcell{8.5}{LLaMA2-7B}
    & FP16 & Baseline & 5.473 & 45.90 & 74.66 & 77.92 & 75.94 & 70.86 & 44.00 & 78.89 & 68.90 & 67.13 \\
    \cmidrule{2-13}
    & \multirowcell{3}{INT2}
    & RTN & 7.8e3 & 26.45 & 26.18 & 39.30 & 25.99 & 0.00 & 24.20 & 49.40 & 49.96 & 30.19 \\
    & & GPTQ & 30.85 & 26.96 & 44.15 & 56.30 & 42.52 & 23.95 & 28.20 & 63.22 & 54.70 & 42.50 \\
    & & \bf{\boa} & \bf{12.76} & 28.33 & 47.73 & 64.71 & 45.46 & 31.33 & 29.60 & 64.85 & 54.46 & \bf{45.81} \\
    \cmidrule{2-13}
    & \multirowcell{3}{INT3}
    & RTN & 342.4 & 22.53 & 35.06 & 44.13 & 28.87 & 1.63 & 26.40 & 58.11 & 48.70 & 33.18 \\
    & & GPTQ & 6.719 & 36.95 & 59.72 & 65.54 & 66.68 & 58.21 & 39.20 & 74.65 & 66.06 & 58.38 \\
    & & \bf{\boa} & \bf{6.007 }& 40.27 & 69.28 & 74.22 & 72.00 & 66.60 & 41.20 & 78.35 & 67.64 & \bf{63.70} \\
    \midrule
    \multirowcell{8.5}{LLaMA2-13B}
    & FP16 & Baseline & 4.885 & 49.06 & 77.65 & 80.49 & 79.38 & 73.41 & 45.80 & 80.69 & 72.22 & 69.84 \\
    \cmidrule{2-13}
    & \multirowcell{3}{INT2}
    & RTN & 5.7e3 & 27.47 & 26.89 & 37.95 & 25.97 & 0.00 & 25.20 & 49.08 & 48.38 & 30.12 \\
    & & GPTQ & 35.08 & 21.76 & 35.31 & 61.96 & 35.22 & 19.41 & 28.80 & 57.29 & 52.88 & 39.08 \\
    & & \bf{\boa} & \bf{18.33} & 29.78 & 46.30 & 62.23 & 49.30 & 29.33 & 27.60 & 63.33 & 52.64 & \bf{45.06} \\
    \cmidrule{2-13}
    & \multirowcell{3}{INT3}
    & RTN & 227.2 & 23.98 & 28.75 & 49.24 & 28.51 & 4.46 & 24.40 & 53.26 & 50.75 & 32.92 \\
    & & GPTQ & 9.790 & 34.81 & 62.12 & 67.06 & 55.57 & 47.61 & 37.00 & 73.18 & 61.33 & 54.84 \\
    & & \bf{\boa} & \bf{5.833 }& 43.52 & 69.28 & 78.93 & 74.71 & 65.22 & 35.20 & 77.42 & 62.51 & \bf{63.35} \\
    \midrule
    \multirowcell{8.5}{LLaMA3-8B}
    & FP16 & Baseline & 6.137 & 53.67 & 77.61 & 81.19 & 79.15 & 72.23 & 45.00 & 81.01 & 73.24 & 70.39 \\
    \cmidrule{2-13}
    & \multirowcell{3}{INT2}
    & RTN & 6.6e4 & 25.51 & 25.80 & 53.94 & 26.34 & 0.00 & 29.00 & 51.52 & 50.20 & 32.79 \\
    & & GPTQ & 24.54 & 23.29 & 32.58 & 53.39 & 39.17 & 7.10 & 27.00 & 53.32 & 52.25 & 36.01 \\
    & & \bf{\boa} & \bf{21.70} & 26.62 & 44.87 & 60.52 & 43.34 & 16.89 & 29.40 & 59.85 & 55.80 & \bf{42.16} \\
    \cmidrule{2-13}
    & \multirowcell{3}{INT3}
    & RTN & 129.1 & 23.21 & 33.88 & 55.60 & 34.83 & 4.60 & 25.20 & 58.22 & 52.57 & 36.01 \\
    & & GPTQ & 8.226 & 41.47 & 63.01 & 75.47 & 70.05 & 59.53 & 40.60 & 73.78 & 69.85 & 61.72 \\
    & & \bf{\boa} & \bf{7.782 }& 45.14 & 72.77 & 78.69 & 72.66 & 62.41 & 42.60 & 77.31 & 71.35 & \bf{65.37} \\
    \midrule
    \multirowcell{8.5}{LLaMA3.2-1B}
    & FP16 & Baseline & 13.15 & 38.14 & 63.26 & 69.51 & 60.78 & 54.38 & 34.60 & 74.37 & 59.51 & 56.82 \\
    \cmidrule{2-13}
    & \multirowcell{3}{INT2}
    & RTN & 6.3e4 & 26.96 & 25.59 & 41.53 & 26.05 & 0.01 & 26.40 & 51.52 & 50.59 & 31.08 \\
    & & GPTQ & 538.9 & 25.26 & 26.64 & 37.83 & 26.41 & 0.22 & 27.60 & 51.41 & 48.46 & 30.48 \\
    & & \bf{\boa} & \bf{312.2} & 25.09 & 26.85 & 40.06 & 27.17 & 1.42 & 27.00 & 51.96 & 51.07 & \bf{31.33} \\
    \cmidrule{2-13}
    & \multirowcell{3}{INT3}
    & RTN & 1.9e3 & 25.60 & 26.94 & 54.13 & 29.90 & 0.58 & 27.00 & 52.18 & 49.17 & 33.19 \\
    & & GPTQ & 112.0 & 24.06 & 39.48 & 53.85 & 31.07 & 13.85 & 27.80 & 60.07 & 49.33 & 37.44 \\
    & & \bf{\boa} & \bf{26.43} & 30.63 & 55.26 & 59.97 & 48.33 & 31.95 & 29.60 & 66.65 & 53.99 & \bf{47.05} \\
    \midrule
    \multirowcell{8.5}{LLaMA3.2-3B}
    & FP16 & Baseline & 11.04 & 46.16 & 67.80 & 78.62 & 70.44 & 62.15 & 36.00 & 75.52 & 67.40 & 63.01 \\
    \cmidrule{2-13}
    & \multirowcell{3}{INT2}
    & RTN & 2.0e4 & 26.79 & 26.52 & 37.89 & 25.93 & 0.00 & 30.80 & 50.92 & 49.09 & 30.99 \\
    & & GPTQ & 98.19 & 24.83 & 27.78 & 52.32 & 33.83 & 4.38 & 28.60 & 52.23 & 51.14 & 34.39 \\
    & & \bf{\boa} & \bf{54.64} & 25.77 & 35.48 & 57.52 & 35.63 & 14.42 & 29.00 & 56.96 & 53.43 & \bf{38.53} \\
    \cmidrule{2-13}
    & \multirowcell{3}{INT3}
    & RTN & 882.6 & 26.37 & 27.86 & 45.87 & 37.04 & 1.44 & 26.00 & 53.92 & 48.46 & 33.37 \\
    & & GPTQ & 46.14 & 28.92 & 37.63 & 44.01 & 39.27 & 18.76 & 28.60 & 61.64 & 54.70 & 39.19 \\
    & & \bf{\boa} & \bf{13.64} & 42.32 & 66.12 & 77.52 & 64.46 & 54.18 & 35.20 & 72.69 & 62.51 & \bf{59.38} \\
    \bottomrule
    \end{tabular}
    \begin{tablenotes}
        \item[*] Results for high bit-widths and results on LLaMA1 and OPT models are provided in \cref{appendix:weight_only_quant_without_transformation} due to the page limitation.
    \end{tablenotes}
    \end{threeparttable}
    \label{tab:weight_only_quant_without_transform_main}

    \vspace{-.3cm}
    
\end{table*}

We measure quantization processing times of \boa \ with and without simultaneous quantization to evaluate how much the head-wise joint quantization can accelerate the quantization process.
From \cref{tab:processing_time_ablation}, we observe that \boa \ requires a significantly long processing time without the simultaneous quantization (more than one day even for the 7B model).
This is because all rows need to be quantized sequentially (\eg, 4096 rows are quantized sequentially for LLaMA-7B) and thus the massive compute capabilities of modern GPUs cannot be utilized properly.
As evident, we can significantly reduce the processing time by applying the head-wise joint quantization (more than 40 times reduction on the 30B model).

\section{Experiments}  \label{sec:experiments}

In this section, we evaluate the weight-only quantization performance of the proposed \boa \ (\cref{subsec:weight-only results}) and synergy with transformation-based methods in terms of weight-only and weight-activation quantization (Sections~\ref{subsec:weight-only results} and~\ref{subsec:weight-activation results}).

\subsection{Experimental Setup}  \label{subsec:setup}

We conduct experiments on OPT~\citep{zhang2022opt}, LLaMA~\citep{touvron2023llama}), LLaMA2~\citep{touvron2023llama2}, and LLaMA3.
As in previous studies~\citep{shao2023omniquant, ma2024affinequant, lin2024duquant, ashkboos2024quarot, liu2024spinquant}, we construct a calibration dataset by sampling 128 random sequences of length 2048 from WikiText-2~\citep{wiki}.
As a performance metric, we use the perplexity (PPL) score on the WikiText-2 test dataset and accuracy on eight zero-shot commonsense reasoning tasks (ARC-challenge (Arc-c) and ARC-easy (Arc-e)~\citep{allenai:arc}, BoolQ (BQ)~\citep{clark2019boolq}, HellaSwag (HS)~\citep{zellers2019hellaswag}, LAMBADA (LAMB)~\citep{paperno2016lambada}, OpenbookQA (OBQA)~\citep{mihaylov2018openbookqa}, PIQA~\citep{bisk2020piqa}, and WinoGrande (WG)~\citep{sakaguchi2021winogrande}).
All experiments were conducted using a single NVIDIA H100 GPU (80~GB).

When determining a quantization order in \boa, the heuristic introduced by GPTQ can be used; the column/row corresponding to the largest $\text{diag}(\mathbf{H}_{\text{col}}) / \text{diag}(\mathbf{H}_{\text{row}})$ (\ie, the most quantization-sensitive column/row) is first quantized for better compensation.
Empirically, we observed that this heuristic could occasionally enhance the performance, yet at other times, it may result in inferior performance.
For GPTQ and the proposed \boa, we conduct experiments with and without this heuristic and report the better results.

\subsection{Weight-Only Quantization Results} \label{subsec:weight-only results}

\paragraph{Comparison with GPTQ} \label{para:weight-rounding performance}

We first compare the proposed \boa \ with GPTQ to demonstrate the efficacy of the proposed attention-aware Hessians.
In this experiment, we do not apply any model transformation method (\eg \ QuaRot~\cite{ashkboos2024quarot} and SpinQuant~\cite{liu2024spinquant}) to solely evaluate the effectiveness of the proposed Hessians.

In \cref{tab:weight_only_quant_without_transform_main} and Tables~\ref{tab:weight_only_quant_without_transform_int4}-\ref{tab:weight_only_quant_without_transform_opt} (see \cref{appendix:weight_only_quant_without_transformation}), we summarize the PPL and zero-shot task performances of \boa \ and GPTQ.
We also include the performance of the rounding-to-nearest (RTN) method which naively assigns the nearest quantization bin.
We observe \boa \ and GPTQ exhibit reasonable PPL even for INT2 quantization where RTN collapses significantly (\ie, PPL is almost $10^{4}$).
This is because \boa \ and GPTQ minimize the task loss degradation rather than the weight quantization error $\Delta \mathbf{W}$.
As evident, \boa \ significantly surpasses GPTQ on all models in both PPL and zero-shot accuracy.
For example, \boa \ achieves 10\%p accuracy improvement on the 3-bit quantized LLaMA2-13B and LLaMA3.2-1B models.
In addition, \boa \ shows 20\%p higher accuracy for the 3-bit quantized LLaMA3.2-3B model.
The key factor leading to such an outstanding performance is that \boa \ considers inter-layer dependencies within the attention module by exploiting the proposed attention-aware Hessians while GPTQ assumes the independence of layers.
\vspace{-1.2mm}

\begin{table}[!t]
    \renewcommand{\arraystretch}{1.0}
    \scriptsize
    \centering
    \vspace{-.25cm}
    \caption{Memory costs of GPTQ and the proposed \boa}
    \begin{threeparttable}
    \begin{tabular}{c c c c c c c c}
    \toprule
    \multirowcell{2}{Method} & \multicolumn{3}{c}{Memory Cost (GB)} & & \multicolumn{3}{c}{Wiki2 PPL ($\downarrow$)} \\
    \cline{2-4} \cline{6-8}
    & 7B & 13B & 30B & & 7B & 13B & 30B \\
    \toprule
    GPTQ & 4.426 & 5.872 & 8.271 & & 19.63 & 34.63 & 9.770 \\
    \midrule
    \boa & 9.550 & 16.56 & 32.79 & & 10.28 & 8.309 & 6.669 \\
    Relaxed \boa$^{*}$ & 6.446 & 8.397 & 11.55 & & 10.53 & 8.700 & 7.007 \\
    \bottomrule
    \end{tabular}
    \begin{tablenotes}
        \item[*] Attention-aware Hessians have been applied to query and key, but not to value.
    \end{tablenotes}
    \end{threeparttable}
    \label{tab:memory_comparison_with_gptq}
    \vspace{-.25cm}
\end{table}

In \cref{tab:memory_comparison_with_gptq}, we summarize the memory costs of \boa \ and GPTQ.
We observe that \boa \ requires larger memory because \boa \ additionally uses outputs of other layers to consider inter-layer dependencies.
It is worth mentioning that when the memory resource is limited, \boa \ can still be used with a slight relaxation.
Noting that the large memory cost of \boa \ is attributable to the Hessian for the value projection ($\mathbf{H}_{\text{col}, h}=2\mathbf{X}\mathbf{A}_{h}^{T} \mathbf{A}_{h} \mathbf{X}^{T}$ in~\cref{eq:hessian_value}) whose shape is $H \times d \times d$, we can greatly reduce the memory cost by exploiting the standard Hessian in~\cref{eq:hessian_optq} for the value projection and applying the proposed attention-aware Hessians only for query and key projections.
In doing so, \boa \ needs slightly more memory (\eg \ 3~GB for 30B; see~\cref{tab:memory_comparison_with_gptq}) yet still performs much better than GPTQ.

We now compare the processing times of \boa \ and GPTQ. 
As evident from~\cref{tab:processing_time}, \boa \ requires a longer processing time than that needed by GPTQ.
This is because GPTQ quantizes all the rows simultaneously, while \boa \ sequentially quantizes them to compensate for the quantization error of each row (see~\cref{fig:boa}(b)).
Clearly, there exists a trade-off between quantization speed and accuracy.
In real cases, when one needs to consider inter-layer dependencies to preserve the performance of the original model as much as possible, the proposed \boa \ would be an intriguing solution when compared to existing gradient-based approaches (see~\cref{tab:processing_time} and~\cref{tab:weight_only_quant_with_transform_main}).
When faster quantization is required, one possible solution is to reduce the number of sequential quantizations by quantizing multiple rows in each head simultaneously, which will be considered in our future studies.

\begin{table}[!t]
    \renewcommand{\arraystretch}{1.0}
    \scriptsize
    \centering
    \vspace{-.25cm}
    \caption{Processing time (hour) of different quantization methods}
    \begin{threeparttable}
    \begin{tabular}{c c c c c c c}
    \toprule
    \multirowcell{2}{Method} & \multirowcell{2}{Inter-layer \\ Dependency} & \multirowcell{2}{Optimization \\ Type} & \multicolumn{3}{c}{LLaMA Model Size} \\
    \cline{4-6}
    & & & 7B & 13B & 30B \\
    \toprule
    GPTQ & X & one-shot & 0.12 & 0.20 & 0.43 \\
    \midrule
    OmniQuant & \multirowcell{3}{O} & \multirowcell{3}{gradient\\-based} & 1.83 & 3.32 & 7.66 \\
    AffineQuant & & & 4.32 & 8.41 & 21.4 \\
    DuQuant+LWC & & & 1.22 & 2.08 & 4.55 \\
    \midrule
    {\bf{\boa}} & O & one-shot & \bf{0.96} & \bf{1.55} & \bf{3.30} \\
    \bottomrule
    \end{tabular}
    \end{threeparttable}
    \label{tab:processing_time}
    \vspace{-.25cm}
\end{table}

\begin{table*}[ht]
    \renewcommand{\arraystretch}{1.0}
    \scriptsize
    \centering
    \caption{Weight-only quantization performance on transformed LLaMA2 and LLaMA3 models}
    \begin{threeparttable}
    \begin{tabular}{c c c c c c c c c c c c c}
    \toprule
    \multirowcell{2}{Model} & \multirowcell{2}{Precision} & \multirowcell{2}{Method} & \multirowcell{2}{Wiki2 PPL ($\downarrow$)} & \multicolumn{9}{c}{Zero-shot Accuracy ($\uparrow$)} \\
    \cline{5-13}
    & & & & Arc-c & Arc-e & BQ & HS & LAMB & OBQA & PIQA & WG & Average \\
    \toprule
    \multirowcell{9.5}{LLaMA2-7B}
    & FP & Baseline & 5.473 & 45.90 & 74.66 & 77.92 & 75.94 & 70.86 & 44.00 & 78.89 & 68.90 & 67.13 \\
    \cmidrule{2-13}
    & \multirowcell{7}{INT2}
    & OmniQuant & 21.85 & 25.17 & 36.91 & 61.80 & 38.38 & 5.80 & 28.40 & 57.40 & 49.49 & 37.92 \\
    &  & AffineQuant & 91.19 & 23.46 & 29.17 & 40.03 & 27.44 & 0.16 & 22.20 & 54.13 & 51.78 & 31.05 \\
    &  & QuaRot-RTN & 1.1e4 & 27.22 & 26.05 & 37.83 & 26.19 & 0.01 & 25.60 & 50.54 & 48.70 & 30.27 \\
    &  & QuaRot-GPTQ & 22.05 & 22.27 & 36.24 & 61.99 & 33.54 & 20.63 & 28.60 & 56.75 & 51.78 & 38.98 \\
    &  & DuQuant & 1.6e4 & 26.62 & 25.72 & 39.08 & 26.06 & 0.00 & 22.20 & 49.73 & 50.12 & 29.94 \\
    &  & DuQuant + LWC & 46.27 & 23.12 & 28.96 & 44.04 & 29.60 & 2.41 & 27.40 & 53.70 & 49.17 & 32.30 \\
    &  & \bf{\boa}$^{\dagger}$ & \bf{10.42} & 29.69 & 54.84 & 64.37 & 52.03 & 46.54 & 33.60 & 67.52 & 59.43 & \bf{51.00} \\
    \midrule
    \multirowcell{9.5}{LLaMA2-13B}
    & FP & Baseline & 4.885 & 49.06 & 77.65 & 80.49 & 79.38 & 73.41 & 45.80 & 80.69 & 72.22 & 69.84 \\
    \cmidrule{2-13}
    & \multirowcell{7}{INT2}
    & OmniQuant & 12.92 & 26.71 & 46.00 & 63.64 & 51.19 & 16.93 & 31.40 & 64.64 & 52.64 & 44.14 \\
    &  & AffineQuant & 9.415 & 26.96 & 49.83 & 63.15 & 52.40 & 26.58 & 33.60 & 64.47 & 53.83 & 46.35 \\
    &  & QuaRot-RTN & 7.9e3 & 27.22 & 26.26 & 37.83 & 25.74 & 0.00 & 24.00 & 49.02 & 49.17 & 29.91 \\
    &  & QuaRot-GPTQ & 9.593 & 31.66 & 56.52 & 63.15 & 48.81 & 36.93 & 32.60 & 66.05 & 60.38 & 49.51 \\
    &  & DuQuant & 1.4e4 & 28.33 & 26.64 & 44.43 & 26.12 & 0.00 & 25.00 & 50.16 & 49.57 & 31.28 \\
    &  & DuQuant + LWC & 10.40 & 28.50 & 46.51 & 63.70 & 52.79 & 25.24 & 30.80 & 65.13 & 54.14 & 45.85 \\
    &  & \bf{\boa}$^{\dagger}$ & \bf{8.237} & 35.49 & 63.97 & 71.28 & 58.36 & 56.25 & 35.40 & 71.82 & 62.75 & \bf{56.92} \\
    \midrule
    \multirowcell{9.5}{LLaMA3-8B}
    & FP & Baseline & 6.137 & 53.67 & 77.61 & 81.19 & 79.15 & 72.23 & 45.00 & 81.01 & 73.24 & 70.39 \\
    \cmidrule{2-13}
    & \multirowcell{7}{INT2}
    & OmniQuant$^{*}$ & 955.8 & 22.78 & 28.24 & 37.83 & 26.18 & 0.00 & 27.20 & 52.83 & 49.01 & 30.51 \\
    &  & AffineQuant$^{*}$ & 1.1e3 & 23.46 & 27.31 & 37.86 & 26.04 & 0.00 & 26.80 & 52.12 & 51.22 & 30.60 \\
    &  & QuaRot-RTN & 3.5e5 & 26.19 & 25.21 & 39.02 & 26.86 & 0.00 & 28.60 & 50.38 & 49.41 & 30.71 \\
    &  & QuaRot-GPTQ & 18.28 & 28.33 & 46.68 & 64.10 & 45.04 & 29.49 & 29.20 & 62.08 & 55.25 & 45.02 \\
    &  & DuQuant & 1.4e6 & 25.00 & 25.34 & 47.37 & 26.68 & 0.00 & 29.80 & 51.96 & 48.93 & 31.89 \\
    &  & DuQuant + LWC & 2.6e4 & 24.40 & 26.68 & 38.26 & 25.30 & 0.00 & 29.20 & 49.67 & 52.01 & 30.69 \\
    &  & \bf{\boa}$^{\dagger}$ & \bf{15.24} & 30.38 & 54.42 & 68.17 & 49.17 & 39.89 & 34.20 & 65.94 & 60.14 & \bf{50.29} \\
    \midrule
    \multirowcell{9.5}{LLaMA3.2-1B}
    & FP & Baseline & 13.15 & 38.14 & 63.26 & 69.51 & 60.78 & 54.38 & 34.60 & 74.37 & 59.51 & 56.82 \\
    \cmidrule{2-13}
    & \multirowcell{7}{INT2}
    & OmniQuant$^{*}$ & 302.3 & 22.01 & 31.69 & 37.95 & 27.64 & 0.14 & 25.60 & 54.95 & 50.12 & 31.26 \\
    &  & AffineQuant$^{*}$ & 268.3 & 22.44 & 31.61 & 37.83 & 27.75 & 0.12 & 26.00 & 54.30 & 49.88 & 31.24 \\
    &  & QuaRot-RTN & 2.6e5 & 26.62 & 25.84 & 38.75 & 26.75 & 0.00 & 28.20 & 51.09 & 51.07 & 31.04 \\
    &  & QuaRot-GPTQ & 54.28 & 22.18 & 34.01 & 56.39 & 31.90 & 14.65 & 24.80 & 56.91 & 50.59 & 36.43 \\
    &  & DuQuant & 4.9e4 & 25.51 & 25.97 & 39.94 & 26.17 & 0.00 & 29.80 & 48.97 & 49.64 & 30.75 \\
    &  & DuQuant + LWC & 9.3e3 & 28.16 & 25.38 & 37.89 & 25.27 & 0.00 & 26.60 & 50.49 & 49.57 & 30.42 \\
    &  & \bf{\boa}$^{\dagger}$ & \bf{40.86} & 24.06 & 39.77 & 58.20 & 34.62 & 16.79 & 26.20 & 57.07 & 52.64 & \bf{38.67} \\
    \midrule
    \multirowcell{9.5}{LLaMA3.2-3B}
    & FP & Baseline & 11.04 & 46.16 & 67.80 & 78.62 & 70.44 & 62.15 & 36.00 & 75.52 & 67.40 & 63.01 \\
    \cmidrule{2-13}
    & \multirowcell{7}{INT2}
    & OmniQuant$^{*}$ & 273.4 & 23.38 & 31.82 & 56.42 & 29.35 & 0.28 & 27.80 & 56.15 & 51.07 & 34.53 \\
    &  & AffineQuant$^{*}$ & 282.2 & 22.35 & 32.87 & 47.28 & 29.44 & 0.25 & 28.20 & 56.26 & 48.38 & 33.13 \\
    &  & QuaRot-RTN & 2.3e4 & 26.19 & 24.49 & 47.09 & 26.55 & 0.00 & 30.40 & 51.74 & 48.30 & 31.85 \\
    &  & QuaRot-GPTQ & 52.18 & 23.98 & 36.95 & 60.92 & 34.68 & 19.24 & 27.40 & 57.67 & 52.49 & 39.17 \\
    &  & DuQuant & 7.7e4 & 25.17 & 25.55 & 41.10 & 26.01 & 0.00 & 28.40 & 51.47 & 50.28 & 31.00 \\
    &  & DuQuant + LWC & 770.9 & 23.63 & 26.64 & 38.20 & 26.32 & 0.03 & 26.20 & 51.74 & 51.70 & 30.56 \\
    &  & \bf{\boa}$^{\dagger}$ & \bf{33.40} & 27.30 & 45.66 & 65.87 & 40.86 & 24.57 & 29.00 & 61.37 & 56.27 & \bf{43.86} \\
    \bottomrule
    \end{tabular}
    \begin{tablenotes}
        \item[$\dagger$] \boa \ has been applied after transforming the model via QuaRot.
        \item[*] The learnable equivalent transformation (LET) option has been deactivated because this option does not support models exploiting grouped-query attention (GQA).
        \item[**] Results for high bit-widths and results on LLaMA1 models are provided in \cref{appendix:weight_only_quant_with_transformation} due to the page limitation.
    \end{tablenotes}
    \end{threeparttable}
    \label{tab:weight_only_quant_with_transform_main}
    \vspace{-.25cm}
\end{table*}

\paragraph{Comparison with transformation-based methods}

To improve the quantization performance, recent studies have transformed models by applying smoothing, rotation, or permutation~\citep{shao2023omniquant,ma2024affinequant,ashkboos2024quarot,liu2024spinquant,lin2024duquant}.
In this experiment, we evaluate \boa \ under those model transformations (see \cref{tab:weight_only_quant_with_transform_main} and Tables~\ref{tab:weight_only_quant_with_transform_int3}-\ref{tab:weight_only_quant_with_transform_llama} in \cref{appendix:weight_only_quant_with_transformation}).
For conventional methods, we ran the official codes provided by the authors and reported the obtained results;
details for implementing conventional methods are provided in \cref{appendix:weight_only_quant_with_transformation}.
When measuring the performance of \boa, we also transformed the model for a fair comparison.
We applied QuaRot for the transformation because QuaRot does not require any training and incurs no extra costs during the actual inference~\citep{ashkboos2024quarot}.

We observe that the performance of the proposed \boa \ is boosted significantly when applying the transformation.
For example, the PPLs of the 2-bit quantized LLaMA2-13B / LLaMA3-8B models improve from 18.33 / 21.70 to 8.237 / 15.24, respectively (see Tables~\ref{tab:weight_only_quant_without_transform_main} and~\ref{tab:weight_only_quant_with_transform_main}).
For INT3 quantization, \boa \ almost preserves the performance of the original full-precision model; the accuracy drop is 2.3\%p for LLaMA3-8B and 1.3\%p for LLaMA2-13B (see~\cref{tab:weight_only_quant_with_transform_int3}).
For all models, we achieve at least 5\%p improvement in the zero-shot accuracy (in particular 11\%p improvement for the 2-bit quantized LLaMA2-13B model).
We note that the performance gap between \boa \ and GPTQ remains significant after the model transformation; for example, the zero-shot accuracy of \boa \ on the 2-bit quantized LLaMA2-7B model is 12\%p larger than that obtained by QuaRot-GPTQ.

Finally, we emphasize that \boa \ not only surpasses existing transformation-based methods exploiting the naive nearest rounding (\ie, OmniQuant, AffineQuant, and DuQuant + LWC) but also is much faster than those methods relying on the gradient-based optimization (see \cref{tab:processing_time}).
For example, on the LLaMA3-8B model, \boa \ achieves 20\%p accuracy improvement for INT2 (see \cref{tab:weight_only_quant_with_transform_main}) and 13\%p improvement for INT3 (see \cref{tab:weight_only_quant_with_transform_int3}) over OmniQuant, AffineQuant, and DuQuant + LWC, yet reduces the quantization processing time of OmniQuant and AffineQuant more than twofold (see \cref{tab:processing_time}).
Moreover, the degree of reduction increases with the model size; \boa \ requires 7 times shorter processing time than that needed by AffineQuant on LLaMA-30B.

\begin{table*}[t]
    \renewcommand{\arraystretch}{1.0}
    \scriptsize
    \centering
    \caption{Weight-activation quantization performance on transformed LLaMA2 and LLaMA3 models}
    \begin{threeparttable}
    \begin{tabular}{c c c c c c c c c c c c c}
    \toprule
    \multirowcell{2}{Model} & \multirowcell{2}{Precision} & \multirowcell{2}{Method} & \multirowcell{2}{Wiki2 PPL ($\downarrow$)} & \multicolumn{9}{c}{Zero-shot Accuracy ($\uparrow$)} \\
    \cline{5-13}
    & & & & Arc-c & Arc-e & BQ & HS & LAMB & OBQA & PIQA & WG & Average \\
    \toprule
    \multirowcell{9}{LLaMA2-7B}
     & FP16 & Baseline & 5.473 & 45.90 & 74.66 & 77.92 & 75.94 & 70.86 & 44.00 & 78.89 & 68.90 & 67.13 \\
     \cmidrule{2-13}
     & \multirowcell{7}{W2A4KV4} 
     & OmniQuant & 1.0e5 & 26.88 & 26.30 & 39.02 & 25.57 & 0.00 & 25.60 & 48.42 & 50.51 & 30.29 \\
     &  & AffineQuant & NaN & NaN & NaN & NaN & NaN & NaN & NaN & NaN & NaN & NaN \\
     &  & SpinQuant-RTN & 23.23 & 25.51 & 34.01 & 62.20 & 32.09 & 14.48 & 26.00 & 55.17 & 50.91 & 37.55 \\
     &  & SpinQuant-GPTQ & 24.29 & 22.95 & 36.74 & 59.88 & 32.83 & 13.81 & 26.20 & 56.64 & 51.30 & 37.54 \\
     &  & DuQuant & 6.4e3 & 28.41 & 26.73 & 38.01 & 25.65 & 0.00 & 26.00 & 48.48 & 51.62 & 30.61 \\
     &  & DuQuant + LWC & 16.35 & 24.91 & 37.33 & 61.99 & 41.46 & 10.54 & 28.20 & 58.71 & 53.28 & 39.55 \\
     &  & \bf{\boa}$^{\dagger}$ & \bf{11.80} & 26.79 & 49.20 & 63.09 & 48.05 & 37.76 & 30.80 & 63.55 & 57.85 & \bf{47.14} \\
    \midrule
    \multirowcell{9}{LLaMA2-13B}
     & FP16 & Baseline & 4.885 & 49.06 & 77.65 & 80.49 & 79.38 & 73.41 & 45.80 & 80.69 & 72.22 & 69.84 \\
     \cmidrule{2-13}
     & \multirowcell{7}{W2A4KV4} 
     & OmniQuant & 3.8e3 & 26.88 & 26.30 & 37.83 & 25.48 & 0.00 & 23.60 & 48.20 & 49.96 & 29.78 \\
     &  & AffineQuant & 1.2e3 & 26.62 & 27.19 & 37.83 & 26.37 & 0.00 & 24.60 & 48.80 & 49.17 & 30.07 \\
     &  & SpinQuant-RTN & 11.71 & 26.96 & 40.61 & 63.12 & 44.54 & 29.64 & 29.20 & 60.88 & 53.67 & 43.58 \\
     &  & SpinQuant-GPTQ & 15.54 & 23.55 & 41.29 & 62.17 & 35.76 & 16.50 & 29.80 & 59.52 & 51.54 & 40.02 \\
     &  & DuQuant & 6.4e3 & 28.41 & 26.73 & 38.01 & 25.65 & 0.00 & 26.00 & 48.48 & 51.62 & 30.61 \\
     &  & DuQuant + LWC & 16.35 & 24.91 & 37.33 & 61.99 & 41.46 & 10.54 & 28.20 & 58.71 & 53.28 & 39.55 \\
     &  & \bf{\boa}$^{\dagger}$ & \bf{8.974} & 31.74 & 55.98 & 64.80 & 56.22 & 49.18 & 34.40 & 68.44 & 59.27 & \bf{52.50} \\
    \midrule
    \multirowcell{9}{LLaMA3-8B}
     & FP16 & Baseline & 6.137 & 53.67 & 77.61 & 81.19 & 79.15 & 72.23 & 45.00 & 81.01 & 73.24 & 70.39 \\
     \cmidrule{2-13}
     & \multirowcell{7}{W2A4KV4} 
     & OmniQuant$^{*}$ & 3.2e5 & 25.85 & 25.46 & 38.26 & 25.35 & 0.00 & 26.80 & 51.58 & 49.09 & 30.30 \\
     &  & AffineQuant$^{*}$ & NaN & NaN & NaN & NaN & NaN & NaN & NaN & NaN & NaN & NaN \\
     &  & SpinQuant-RTN & 31.27 & 24.40 & 33.50 & 62.14 & 36.01 & 19.99 & 27.40 & 56.69 & 52.80 & 39.12 \\
     &  & SpinQuant-GPTQ & 29.60 & 22.18 & 34.89 & 58.23 & 35.07 & 13.42 & 27.20 & 55.88 & 51.22 & 37.26 \\
     &  & DuQuant & 5.4e5 & 27.30 & 24.62 & 50.67 & 26.49 & 0.00 & 28.20 & 50.82 & 50.59 & 32.34 \\
     &  & DuQuant + LWC & 4.1e4 & 25.60 & 26.09 & 37.86 & 25.55 & 0.00 & 25.60 & 51.36 & 49.25 & 30.16 \\
     &  & \bf{\boa}$^{\dagger}$ & \bf{18.23} & 28.92 & 48.86 & 62.54 & 44.62 & 29.38 & 28.80 & 62.79 & 54.22 & \bf{45.02} \\
    \midrule
    \multirowcell{9}{LLaMA3.2-1B}
     & FP16 & Baseline & 13.15 & 38.14 & 63.26 & 69.51 & 60.78 & 54.38 & 34.60 & 74.37 & 59.51 & 56.82 \\
     \cmidrule{2-13}
     & \multirowcell{7}{W2A4KV4} 
     & OmniQuant$^{*}$ & 7.0e3 & 24.83 & 26.26 & 37.98 & 25.59 & 0.00 & 27.60 & 49.84 & 50.75 & 30.36 \\
     &  & AffineQuant$^{*}$ & NaN & NaN & NaN & NaN & NaN & NaN & NaN & NaN & NaN & NaN \\
     &  & SpinQuant-RTN & 110.6 & 22.61 & 32.07 & 52.14 & 28.59 & 4.08 & 28.20 & 52.83 & 49.49 & 33.75 \\
     &  & SpinQuant-GPTQ & 143.8 & 22.35 & 31.23 & 51.96 & 28.91 & 2.91 & 26.20 & 53.32 & 51.85 & 33.59 \\
     &  & DuQuant & 5.6e4 & 26.62 & 24.71 & 41.74 & 26.16 & 0.01 & 28.20 & 48.80 & 50.51 & 30.84 \\
     &  & DuQuant + LWC & 8.5e3 & 28.07 & 26.01 & 38.26 & 25.84 & 0.00 & 28.80 & 50.11 & 50.91 & 31.00 \\
     &  & \bf{\boa}$^{\dagger}$ & \bf{77.05} & 22.61 & 36.66 & 57.22 & 32.13 & 8.07 & 25.80 & 55.55 & 50.99 & \bf{36.13} \\
    \midrule
    \multirowcell{9}{LLaMA3.2-3B}
     & FP16 & Baseline & 11.04 & 46.16 & 67.80 & 78.62 & 70.44 & 62.15 & 36.00 & 75.52 & 67.40 & 63.01 \\
     \cmidrule{2-13}
     & \multirowcell{7}{W2A4KV4} 
     & OmniQuant$^{*}$ & 6.6e3 & 27.82 & 25.00 & 38.35 & 25.43 & 0.00 & 28.00 & 52.18 & 49.80 & 30.82 \\
     &  & AffineQuant$^{*}$ & 6.0e3 & 25.94 & 26.89 & 38.10 & 25.66 & 0.01 & 28.40 & 51.58 & 49.80 & 30.80 \\
     &  & SpinQuant-RTN & 51.19 & 24.83 & 31.48 & 45.32 & 29.90 & 11.05 & 27.80 & 55.17 & 50.20 & 34.47 \\
     &  & SpinQuant-GPTQ & 65.09 & 23.21 & 33.54 & 38.26 & 31.16 & 7.18 & 25.80 & 55.98 & 51.38 & 33.31 \\
     &  & DuQuant & 1.2e5 & 26.02 & 25.51 & 49.02 & 26.74 & 0.00 & 29.00 & 51.03 & 49.88 & 32.15 \\
     &  & DuQuant + LWC & 1.7e3 & 24.40 & 26.09 & 37.83 & 25.91 & 0.00 & 27.80 & 50.49 & 52.49 & 30.63 \\
     &  & \bf{\boa}$^{\dagger}$ & \bf{37.12} & 27.99 & 37.67 & 59.30 & 37.91 & 15.81 & 27.40 & 57.73 & 52.41 & \bf{39.53} \\
    \bottomrule
    \end{tabular}
    \begin{tablenotes}
        \item[$\dagger$] \boa \ has been applied after transforming the model via SpinQuant.
        \item[*] The LET option has been deactivated because this option does not support models exploiting GQA.
        \item[**] `NaN` means that loss diverges in the quantization process.
        \item[***] Results for other configurations are provided in \cref{appendix:weight_act_quant} due to the page limitation.
    \end{tablenotes}
    \end{threeparttable}
    \label{tab:weight_act_quant_main}
    \vspace{-.25cm}
\end{table*}

\subsection{Weight-Activation Quantization Results} \label{subsec:weight-activation results}

We now evaluate the weight-activation quantization performance (see  \cref{tab:weight_act_quant_main} and Tables~\ref{tab:weight_act_quant_llama3-8b}-\ref{tab:weight_act_quant_llama3.2-3b} in \cref{appendix:weight_act_quant}).
As in previous studies~\cite{shao2023omniquant, ma2024affinequant, lin2024duquant, ashkboos2024quarot, liu2024spinquant}, we quantize input activations to all linear layers and KV caches with the Min-Max nearest-rounding quantizer where quantization parameters are determined dynamically for each token.
We use the notation `WxAyKVz' to denote the x-bit weight quantization, y-bit input activation quantization, and z-bit KV cache quantization.
In this experiment, when measuring the performance of RTN, GPTQ, and the proposed \boa, we use SpinQuant (instead of QuaRot) for the model transformation.
It is worth noting that while QuaRot uses Hadamard matrices (that are independent of data) for the rotation~\cite{ashkboos2024quarot}, SpinQuant optimizes rotation matrices that make models more robust to the activation quantization and thus could outperform QuaRot~\cite{liu2024spinquant}.
In our experiments, rotation matrices have been optimized with respect to activation-only quantized networks (\ie, weights are fixed with full-precision) as in~\cite{liu2024spinquant}.

From \cref{tab:weight_act_quant_main} and Tables~\ref{tab:weight_act_quant_llama3-8b}-\ref{tab:weight_act_quant_llama3.2-3b} in \cref{appendix:weight_act_quant}, we observe that the outstanding weight-quantization performance of the proposed \boa \ leads to the state-of-the-art performance for the weight-activation quantization.
In particular, the performance gain is noticeable for low-bit (\eg, W2A4KV4 and W2A4KV16). 
For example, compared to SpinQuant-GPTQ, \boa \ achieves 10\%p accuracy improvement on LLaMA2-7B and 12.5\%p improvement on LLaMA2-13B (see \cref{tab:weight_act_quant_main}). 
Compared to the conventional gradient-based methods (\ie \ OmniQuant, AffineQuant, and DuQuant), \boa \ achieves 8\%p, 13\%p, 13\%p, 5\%p, and 7\%p improvement on LLaMA2-7B, LLaMA2-13B, LLaMA3-8B, LLaMA3.2-1B, and LLaMA3.2-3B, respectively.

\section{Conclusion}  \label{sec:conclusion}

In this paper, we proposed a novel backpropagation-free PTQ algorithm called \boa.
To consider the inter-layer dependencies within the attention module, we approximated the Hessian matrices by exploiting the attention reconstruction error rather than the layer-wise reconstruction error.
To mitigate the computational overhead incurred by the proposed attention-aware Hessians, we also incorporated several techniques, including Hessian relaxation, efficient computation of inverse and Cholesky decomposition of attention-aware Hessians, and simultaneous quantization of different attention heads.
Finally, from extensive experiments, we demonstrated the efficacy of the proposed \boa.

\newpage

\section*{Acknowledgment}
We would like thanks to Daehyun Kim, Ph.D., Hyeonmok Ko, Ph.D., and Hyemi Jang, Ph.D. for their helpful discussion.

\section*{Impact Statement}

\boa~is a highly efficient and accurate quantization algorithm that minimizes accuracy loss while significantly reducing the time required for model quantization and deployment. By synergizing with other methods, \boa \ facilitates the efficient operation of LLMs on resource-constrained hardware using only integer arithmetic. This breakthrough is particularly impactful for on-device AI, allowing real-time model inference on mobile and edge devices without the need for server access, paving the way for scalable, decentralized AI solutions.





\bibliography{main}
\bibliographystyle{icml2025}

\newpage
\appendix
\onecolumn

\section{Proof of \cref{footnote:hessian}}  \label{appendix:hessian_derivation}

In our proof, we use the following useful properties of the Kronecker product:
\begin{subequations}
\begin{align}
    \vect \left ( \mathbf{M}_{1} \mathbf{M}_{2} \mathbf{M}_{3} \right ) &= \left ( \mathbf{M}_{3}^{T} \otimes \mathbf{M}_{1} \right ) \vect (\mathbf{M}_{2}), \label{eq:Kron-prop_1} \\
    \left ( \mathbf{M}_{1} \otimes \mathbf{M}_{2} \right )^{T} &= \mathbf{M}_{1}^{T} \otimes \mathbf{M}_{2}^{T}, \label{eq:Kron-prop_2} \\
    \left ( \mathbf{M}_{1} \otimes \mathbf{M}_{2} \right ) \left ( \mathbf{M}_{3} \otimes \mathbf{M}_{4} \right ) &= \mathbf{M}_{1} \mathbf{M}_{3} \otimes \mathbf{M}_{2} \mathbf{M}_{4}, \label{eq:Kron-prop_3}
\end{align}
\end{subequations}
where $\vect(\cdot)$ denotes the vectorization operation. 

Using~\cref{eq:Kron-prop_1}, we have
\begin{align*}
    \left \| \mathbf{M}_{1} \Delta \mathbf{W} \mathbf{M}_{2} \right \|_{F}^{2}
        &= \left \| \left ( \mathbf{M}_{2}^{T} \otimes \mathbf{M}_{1} \right ) \Delta \mathbf{w} \right \|_{2}^{2} 
        = \Delta \mathbf{w}^{T} \left ( \mathbf{M}_{2}^{T} \otimes \mathbf{M}_{1} \right )^{T} \left ( \mathbf{M}_{2}^{T} \otimes \mathbf{M}_{1} \right ) \Delta \mathbf{w},
\end{align*}
where $\Delta \mathbf{w} = \vect(\Delta \mathbf{W})$.
In addition, by~\cref{eq:Kron-prop_2} and~\cref{eq:Kron-prop_3}, we have
\begin{align*}
    \Delta \mathbf{w}^{T} \left ( \mathbf{M}_{2}^{T} \otimes \mathbf{M}_{1} \right )^{T} \left ( \mathbf{M}_{2}^{T} \otimes \mathbf{M}_{1} \right ) \Delta \mathbf{w}
        &= \Delta \mathbf{w}^{T} \left ( \mathbf{M}_{2} \otimes \mathbf{M}_{1}^{T} \right ) \left ( \mathbf{M}_{2}^{T} \otimes \mathbf{M}_{1} \right ) \Delta \mathbf{w} \nonumber \\
        &= \Delta \mathbf{w}^{T} \left ( \mathbf{M}_{2} \mathbf{M}_{2}^{T} \otimes \mathbf{M}_{1}^{T} \mathbf{M}_{1} \right ) \Delta \mathbf{w}.
\end{align*}
Finally, by exploiting the fact that $\frac{\partial^{2}\mathbf{x}^{T} \mathbf{A} \mathbf{x}}{\partial \mathbf{x}^{2}} = \mathbf{A} + \mathbf{A}^{T}$, we obtain
\begin{align*}
    \frac{\partial^{2} \left \| \mathbf{M}_{1} \Delta \mathbf{W} \mathbf{M}_{2} \right \|_{F}^{2}}{\partial \Delta \mathbf{w}^{2}}
        &= \mathbf{M}_{2} \mathbf{M}_{2}^{T} \otimes \mathbf{M}_{1}^{T} \mathbf{M}_{1} + \left ( \mathbf{M}_{2} \mathbf{M}_{2}^{T} \otimes \mathbf{M}_{1}^{T} \mathbf{M}_{1} \right )^{T} \\
        &\overset{(a)}{=} \mathbf{M}_{2} \mathbf{M}_{2}^{T} \otimes \mathbf{M}_{1}^{T} \mathbf{M}_{1} + \left ( \mathbf{M}_{2} \mathbf{M}_{2}^{T} \right )^{T} \otimes \left ( \mathbf{M}_{1}^{T} \mathbf{M}_{1} \right )^{T} \\
        &= 2\mathbf{M}_{2} \mathbf{M}_{2}^{T} \otimes \mathbf{M}_{1}^{T} \mathbf{M}_{1},
\end{align*}
where (a) follows from~\cref{eq:Kron-prop_2}.
This completes the proof.

\section{Attention-aware Hessians for Models Exploiting RoPE}  \label{appendix:hessian_derivation_rope}

As mentioned, when RoPE is applied, the proposed surrogate $\| \mathbf{K}_{h} \Delta \mathbf{Q}_{h}^{T} \|_{F}^{2}$ used to develop the attention-aware Hessian $\mathbf{H}^{(\mathbf{w}_{Q, h})}$ in~\cref{eq:relaxed-hessian_query} changes to $\| \widetilde{\mathbf{K}}_{h} \Delta \widetilde{\mathbf{Q}}_{h}^{T} \|_{F}^{2}$, where $\widetilde{\mathbf{K}}_{h} = \rope(\mathbf{K}_{h})$ and $\widetilde{\mathbf{Q}}_{h} = \rope(\mathbf{Q}_{h})$.
Let $\mathbf{R}_{\ell}$ be the rotary matrix for the $\ell$-th token (see eq.~(15) in~\cite{rope}) and $\widetilde{\mathbf{Q}}_{h}^{T} = [\widetilde{\mathbf{q}}_{h, 1} \ \ldots \widetilde{\mathbf{q}}_{h, L}]$, then the objective can be expressed as
$$\| \widetilde{\mathbf{K}}_{h} \Delta \widetilde{\mathbf{Q}}_{h}^{T} \|_{F}^{2} = \sum_{\ell=1}^{L} \| \widetilde{\mathbf{K}}_{h} \Delta \widetilde{\mathbf{q}}_{h, \ell} \|_{2}^{2} = \sum_{\ell=1}^{L} \| \widetilde{\mathbf{K}}_{h} \Delta (\mathbf{R}_{\ell} \mathbf{W}_{Q, h} \mathbf{x}_{\ell}) \|_{2}^{2} = \sum_{\ell=1}^{L} \| \widetilde{\mathbf{K}}_{h} \mathbf{R}_{\ell} \Delta \mathbf{W}_{Q, h} \mathbf{x}_{\ell} \|_{2}^{2},$$
which yields the following Hessian equation (see \cref{footnote:hessian}):
$$\mathbf{H}^{(\mathbf{w}_{Q, h})} = \sum_{\ell=1}^{L} ( 2\mathbf{x}_{\ell} \mathbf{x}_{\ell}^{T} \otimes \mathbf{R}_{\ell}^{T} \widetilde{\mathbf{K}}_{h}^{T} \widetilde{\mathbf{K}}_{h} \mathbf{R}_{\ell} ).$$
Finally, we take the factorized approximation for the summation of Kronecker products (i.e., $\mathbb{E} [\mathbf{M}_{1} \otimes \mathbf{M}_{2}] \approx \mathbb{E} [\mathbf{M}_{1}] \otimes \mathbb{E} [\mathbf{M}_{2}]$; see eq.~(20) in \cite{botev2017practical}):
$$\mathbf{H}^{(\mathbf{w}_{Q, h})} \approx \sum_{\ell=1}^{L} 2\mathbf{x}_{\ell} \mathbf{x}_{\ell}^{T} \otimes \frac{1}{L} \sum_{\ell=1}^{L} \mathbf{R}_{\ell}^{T} \widetilde{\mathbf{K}}_{h}^{T} \widetilde{\mathbf{K}}_{h} \mathbf{R}_{\ell} = 2\mathbf{X} \mathbf{X}^{T} \otimes \frac{1}{L} \sum_{\ell=1}^{L} \mathbf{R}_{\ell}^{T} \widetilde{\mathbf{K}}_{h}^{T} \widetilde{\mathbf{K}}_{h} \mathbf{R}_{\ell}.$$
By taking similar steps, the attention-aware Hessian for the key projection weight $\mathbf{W}_{K, h}$ with RoPE can be established as 
$$\mathbf{H}^{(\mathbf{w}_{K, h})} = 2\mathbf{X} \mathbf{X}^{T} \otimes \frac{1}{L} \sum_{\ell=1}^{L} \mathbf{R}_{\ell}^{T} \widetilde{\mathbf{Q}}_{h}^{T} \widetilde{\mathbf{Q}}_{h} \mathbf{R}_{\ell}.$$

\section{Refined Weight-update Formula}  \label{appendix:refined_weight_update}

We recall that the Hessian-based weight-update formula is given by~\citep{obq, frantar2023optq}
\begin{align*}
    \bm{\delta} \vw &= - \frac{w_{q} - \gQ(w_{q})}{[ \mathbf{U} ]_{q, q}} [ \mathbf{U} ]_{q, :} \text{ where } \mathbf{U} = \chol ( \mathbf{H}^{-1} )^{T}.
\end{align*}

For the proposed attention-aware Hessians in~\cref{tab:hessians}, we have
\begin{align*}
    \mathbf{U}_{h} = \mathbf{U}_{\text{col}, h} \otimes \mathbf{U}_{\text{row}, h},
\end{align*}
where $\mathbf{U}_{\text{col}, h} = \chol (\mathbf{H}_{\text{col}, h}^{-1})^{T}$ and $\mathbf{U}_{\text{row}, h} = \chol (\mathbf{H}_{\text{row}, h}^{-1})^{T}$ (see \cref{subsec:bptq}). 
Therefore, the weight-update formula can be recast as
\begin{align*}
    \bm{\delta} \vw_{h} &= - \frac{w_{q} - \gQ(w_{q})}{[ \mathbf{U}_{\text{col}, h} \otimes \mathbf{U}_{\text{row}, h} ]_{q, q}} [ \mathbf{U}_{\text{col}, h} \otimes \mathbf{U}_{\text{row}, h} ]_{q, :}.
\end{align*}

\begin{figure*}[htb]
    \centering
    \includegraphics[width=.6\linewidth]{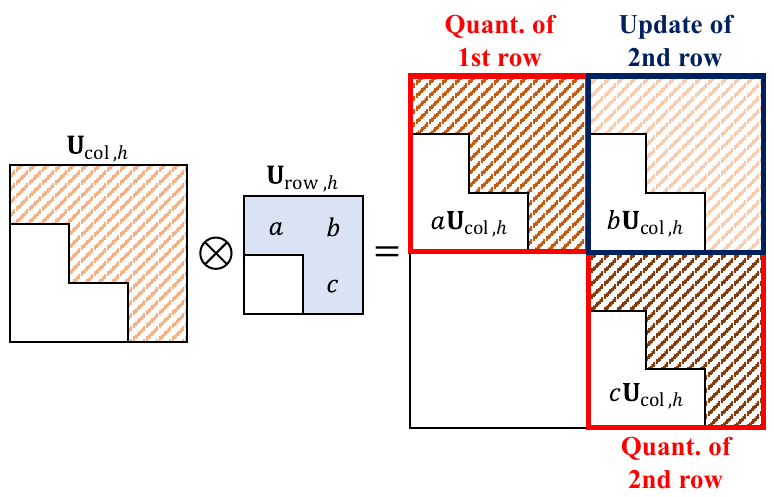}
    \vspace{-0.2cm}
    \caption{Illustration of the Hessian information when $d_{\text{row}}=2$ and $d_{\text{col}}=3$}
    \vspace{-0.2cm}
    \label{fig:hessian-info}
\end{figure*}

For simplicity, suppose we quantize the first ($0$-th) row.
When the weight $[\mathbf{W}_{h}]_{0, j} (= [\mathbf{W}^{(0)}]_{h, j})$ in the $j$-th column is quantized, the weight-update of the $i$-th row is simplified as (see \cref{fig:hessian-info} for the ease of understanding)
\begin{align*}
    [\bm{\delta} \mathbf{W}_{h} ]_{i, :}
        &= - \frac{[\mathbf{W}_{h}]_{0, j} - \gQ([\mathbf{W}_{h}]_{0, j})}{[\mathbf{U}_{\text{row}, h}]_{0, 0} [\mathbf{U}_{\text{col}, h}]_{j, j}} [\mathbf{U}_{\text{row}, h}]_{0, i} [\mathbf{U}_{\text{col}, h}]_{j, :} \\
        &= - \frac{[\mathbf{W}_{h}]_{0, j} - \gQ([\mathbf{W}_{h}]_{0, j})}{[\mathbf{U}_{\text{col}, h}]_{j, j}} \cdot \frac{[\mathbf{U}_{\text{row}, h}]_{0, i} [\mathbf{U}_{\text{col}, h}]_{j, :}}{[\mathbf{U}_{\text{row}, h}]_{0, 0}}
\end{align*}
Thus, after the quantization of all weights in the first row, the total amount of the weight-update for the $i$-th row can be expressed as
\begin{align*}
    [\bm{\delta} \mathbf{W}_{h, \text{total}}]_{i, :} 
        &= -\sum_{j=0}^{d_{\text{col}} - 1} \frac{[\mathbf{W}_{h}]_{0, j} - \gQ([\mathbf{W}_{h}]_{0, j})}{[\mathbf{U}_{\text{col}, h}]_{j, j}} \cdot \frac{[\mathbf{U}_{\text{row}, h}]_{0, i} [\mathbf{U}_{\text{col}, h}]_{j, :}}{[\mathbf{U}_{\text{row}, h}]_{0, 0}} \\
        &= -\frac{[\mathbf{U}_{\text{row}, h}]_{0, i}}{[\mathbf{U}_{\text{row}, h}]_{0, 0}} \sum_{j=0}^{d_{\text{col}} - 1} \frac{[\mathbf{W}_{h}]_{0, j} - \gQ([\mathbf{W}_{h}]_{0, j})}{[\mathbf{U}_{\text{col}, h}]_{j, j}} \cdot [\mathbf{U}_{\text{col}, h}]_{j, :}.
\end{align*}
Furthermore, by noting that (see line~5 in \cref{algo:gptq})
\begin{align*}
    [\mathbf{E}_{\text{GPTQ}}]_{h, j} &= \frac{[\mathbf{W}_{h}]_{0, j} - \gQ([\mathbf{W}_{h}]_{0, j})}{[\mathbf{U}_{\text{col}, h}]_{j, j}},
\end{align*}
we obtain
\begin{align*}
    [\bm{\delta} \mathbf{W}_{h, \text{total}}]_{i, :} 
        &= -\frac{[\mathbf{U}_{\text{row}, h}]_{0, i}}{[\mathbf{U}_{\text{row}, h}]_{0, 0}} \sum_{j=0}^{d_{\text{col}} - 1} [\mathbf{E}_{\text{GPTQ}}]_{h, j} \cdot [\mathbf{U}_{\text{col}, h}]_{j, :}
        = -\frac{[\mathbf{U}_{\text{row}, h}]_{0, i}}{[\mathbf{U}_{\text{row}, h}]_{0, 0}}  [\mathbf{E}_{\text{GPTQ}}]_{h, :} \mathbf{U}_{\text{col}, h}.
\end{align*}
As a result, the weight-update matrix to compensate for the quantization error of the first row is given by
\begin{align}
    [\bm{\delta} \mathbf{W}_{h, \text{total}}]_{0:, :}
        &= -\frac{[\mathbf{U}_{\text{row}, h}^{T}]_{0:, 0} [\mathbf{E}_{\text{GPTQ}}]_{h, :} \mathbf{U}_{\text{col}, h}}{[\mathbf{U}_{\text{row}, h}]_{0, 0}}.  \label{eq:refine_weight_update-first}
\end{align}
By taking similar steps as above, we can easily generalize~\cref{eq:refine_weight_update-first} for the $j$-th row as follows:
\begin{align}
    [\bm{\delta} \mathbf{W}_{h, \text{total}}]_{j:, :}
        &= -\frac{[\mathbf{U}_{\text{row}, h}^{T}]_{j:, j} [\mathbf{E}_{\text{GPTQ}}]_{h, :} \mathbf{U}_{\text{col}, h}}{[\mathbf{U}_{\text{row}, h}]_{j, j}}.  \label{eq:refine_weight_update}
\end{align}

\newpage

\section{Additional Experimental Results}

\subsection{Performance of Weight-Rounding Optimization}  \label{appendix:weight_only_quant_without_transformation}

We evaluate the weight-only quantization performance of the proposed \boa.
To solely compare the performance of the weight-rounding optimization, we do not apply any transform (\eg, smoothing, rotation, and permutation) to the models.
In this appendix, we provide the results for the INT4 weight quantization of LLaMA2 and LLaMA3 models (see \cref{tab:weight_only_quant_without_transform_int4}) and the results on LLaMA1 models (see \cref{tab:weight_only_quant_without_transform_llama}) and OPT models (see \cref{tab:weight_only_quant_without_transform_opt}).

\begin{table*}[!h]
    \renewcommand{\arraystretch}{1.0}
    \tiny
    \centering
    \caption{INT4 weight-only quantization performance on LLaMA2 and LLaMA3 models without transformation}
    \begin{threeparttable}
    \begin{tabular}{c c c c c c c c c c c c c}
    \toprule
    \multirowcell{2}{Model} & \multirowcell{2}{Precision} & \multirowcell{2}{Method} & \multirowcell{2}{Wiki2 PPL ($\downarrow$)} & \multicolumn{9}{c}{Zero-shot Accuracy ($\uparrow$)} \\
    \cline{5-13}
    & & & & Arc-c & Arc-e & BQ & HS & LAMB & OBQA & PIQA & WG & Average \\
    \toprule
    \multirowcell{4.7}{LLaMA2-7B}
    & FP & Baseline & 5.473 & 45.90 & 74.66 & 77.92 & 75.94 & 70.86 & 44.00 & 78.89 & 68.90 & 67.13 \\
    \cmidrule{2-13}
    & \multirowcell{3}{INT4} 
    & RTN & 6.998 & 41.64 & 67.05 & 72.72 & 71.36 & 59.78 & 42.60 & 77.04 & 65.75 & 62.24 \\
    & & GPTQ & 5.809 & 42.06 & 69.82 & 67.06 & 70.44 & 62.80 & 43.00 & 77.15 & 68.11 & 62.56 \\
    & & \bf{\boa} & \bf{5.622} & 43.94 & 72.26 & 78.04 & 74.53 & 68.96 & 44.20 & 78.40 & 69.14 & \bf{66.18} \\
    \midrule
    \multirowcell{4.7}{LLaMA2-13B}
    & FP & Baseline & 4.885 & 49.06 & 77.65 & 80.49 & 79.38 & 73.41 & 45.80 & 80.69 & 72.22 & 69.84 \\
    \cmidrule{2-13}
    & \multirowcell{3}{INT4} 
    & RTN & 6.319 & 43.00 & 66.88 & 71.01 & 66.14 & 62.25 & 38.20 & 77.04 & 63.85 & 61.05 \\
    & & GPTQ & 5.632 & 47.18 & 74.49 & 74.37 & 75.00 & 70.89 & 44.20 & 78.40 & 69.69 & 66.78 \\
    & & \bf{\boa} & \bf{5.096} & 47.95 & 75.80 & 78.93 & 77.96 & 70.61 & 41.60 & 79.87 & 71.35 & \bf{68.01} \\
    \midrule
    \multirowcell{4.7}{LLaMA3-8B}
    & FP & Baseline & 6.137 & 53.67 & 77.61 & 81.19 & 79.15 & 72.23 & 45.00 & 81.01 & 73.24 & 70.39 \\
    \cmidrule{2-13}
    & \multirowcell{3}{INT4} & RTN & 7.981 & 49.06 & 73.65 & 73.30 & 76.62 & 59.55 & 45.00 & 78.29 & 72.93 & 66.05 \\
    & & GPTQ & 7.960 & 49.06 & 73.44 & 73.70 & 69.30 & 60.19 & 45.00 & 79.54 & 73.56 & 65.47 \\
    & & \bf{\boa} & \bf{6.561} & 50.17 & 77.74 & 80.31 & 77.42 & 70.20 & 44.40 & 80.14 & 73.88 & \bf{69.28} \\
    \midrule
    \multirowcell{4.7}{LLaMA3.2-1B}
    & FP & Baseline & 13.15 & 38.14 & 63.26 & 69.51 & 60.78 & 54.38 & 34.60 & 74.37 & 59.51 & 56.82 \\
    \cmidrule{2-13}
    & \multirowcell{3}{INT4} 
    & RTN & 18.81 & 34.30 & 57.41 & 65.20 & 55.93 & 32.72 & 33.00 & 69.15 & 56.35 & 50.51 \\
    & & GPTQ & 16.20 & 35.41 & 58.63 & 64.46 & 55.69 & 38.08 & 32.20 & 69.04 & 56.27 & 51.22 \\
    & & \bf{\boa} & \bf{14.29}  & 36.18 & 60.82 & 67.22 & 59.03 & 51.65 & 34.00 & 71.87 & 58.09 & \bf{54.86} \\
    \midrule
    \multirowcell{4.7}{LLaMA3.2-3B}
    & FP & Baseline & 11.04 & 46.16 & 67.80 & 78.62 & 70.44 & 62.15 & 36.00 & 75.52 & 67.40 & 63.01 \\
    \cmidrule{2-13}
    & \multirowcell{3}{INT4} 
    & RTN & 13.86 & 43.26 & 61.24 & 73.67 & 68.27 & 54.18 & 38.40 & 71.76 & 64.01 & 59.35 \\
    & & GPTQ & 12.41 & 44.28 & 63.51 & 77.00 & 68.44 & 57.39 & 36.60 & 74.16 & 67.01 & 61.05 \\
    & & \bf{\boa} & \bf{11.74}  & 44.11 & 66.75 & 77.86 & 69.42 & 61.74 & 36.20 & 75.08 & 66.85 & \bf{62.25} \\
    \bottomrule
    \end{tabular}
    \end{threeparttable}

    \label{tab:weight_only_quant_without_transform_int4}
    
\end{table*}

\begin{table*}[!h]
    \renewcommand{\arraystretch}{1.0}
    \tiny
    \centering
    
    \caption{Weight-only quantization performance on LLaMA models without transformation}
    \begin{threeparttable}
    \begin{tabular}{c c c c c c c c c c c c c}
    \toprule
    \multirowcell{2}{Model} & \multirowcell{2}{Precision} & \multirowcell{2}{Method} & \multirowcell{2}{Wiki2 PPL ($\downarrow$)} & \multicolumn{9}{c}{Zero-shot Accuracy ($\uparrow$)} \\
    \cline{5-13}
    & & & & Arc-c & Arc-e & BQ & HS & LAMB & OBQA & PIQA & WG & Average \\
    \toprule
    \multirowcell{11.8}{LLaMA-7B}
    & FP & Baseline & 5.677 & 44.71 & 72.94 & 75.05 & 76.21 & 70.66 & 44.40 & 79.16 & 70.01 & 66.64 \\
    \cmidrule{2-13}
    & \multirowcell{3}{INT2} 
    & RTN & 5.9e3 & 26.37 & 26.73 & 39.82 & 26.37 & 0.01 & 26.00 & 51.80 & 48.62 & 30.72 \\
    &  & GPTQ & 19.63 & 25.34 & 41.58 & 56.27 & 37.27 & 26.24 & 30.20 & 61.70 & 55.56 & 41.77 \\
    &  & \bf{\boa} & \bf{10.28} & 28.75 & 52.69 & 64.19 & 49.34 & 41.48 & 32.60 & 66.54 & 59.19 & \bf{49.35} \\
    \cmidrule{2-13}
    & \multirowcell{3}{INT3} 
    & RTN & 61.31 & 26.45 & 50.46 & 59.76 & 40.07 & 14.64 & 33.00 & 67.08 & 55.88 & 43.42 \\
    &  & GPTQ & 18.44 & 33.53 & 61.36 & 64.40 & 56.22 & 44.91 & 35.40 & 72.09 & 64.33 & 54.03 \\
    &  & \bf{\boa} & \bf{6.267} & 39.76 & 67.68 & 75.35 & 71.01 & 68.92 & 41.80 & 77.80 & 67.48 & \bf{63.73} \\
    \cmidrule{2-13}
    & \multirowcell{3}{INT4} 
    & RTN & 7.912 & 41.47 & 70.29 & 72.81 & 73.16 & 63.81 & 41.60 & 77.86 & 68.27 & 63.66 \\
    &  & GPTQ & 7.791 & 40.36 & 66.50 & 70.58 & 68.48 & 61.84 & 39.00 & 77.42 & 68.43 & 61.58 \\
    &  & \bf{\boa} & \bf{5.899} & 43.77 & 72.18 & 75.54 & 74.75 & 69.60 & 44.40 & 78.62 & 69.77 & \bf{66.08} \\
    \midrule
    \multirowcell{11.8}{LLaMA-13B}
    & FP & Baseline & 5.090 & 47.70 & 74.75 & 77.95 & 79.07 & 73.66 & 44.80 & 80.14 & 72.77 & 68.86 \\
    \cmidrule{2-13}
    & \multirowcell{3}{INT2} 
    & RTN & 2.2e3 & 25.60 & 26.01 & 39.48 & 26.00 & 0.00 & 25.20 & 49.46 & 48.54 & 30.04 \\
    &  & GPTQ & 34.63 & 26.37 & 43.43 & 56.67 & 46.85 & 33.11 & 31.40 & 65.07 & 57.62 & 45.07 \\
    &  & \bf{\boa} & \bf{8.309} & 32.76 & 59.51 & 69.57 & 59.52 & 51.79 & 35.20 & 71.22 & 65.19 & \bf{55.60} \\
    \cmidrule{2-13}
    & \multirowcell{3}{INT3} 
    & RTN & 134.5 & 25.00 & 49.24 & 56.39 & 36.91 & 11.41 & 29.20 & 64.20 & 54.62 & 40.87 \\
    &  & GPTQ & 7.096 & 42.24 & 69.74 & 70.73 & 73.40 & 62.99 & 39.20 & 77.09 & 68.11 & 62.94 \\
    &  & \bf{\boa} & \bf{5.461} & 45.82 & 71.93 & 75.96 & 75.71 & 70.62 & 43.40 & 78.35 & 70.24 & \bf{66.50} \\
    \cmidrule{2-13}
    & \multirowcell{3}{INT4} 
    & RTN & 7.742 & 45.22 & 69.99 & 72.42 & 75.67 & 65.65 & 41.60 & 78.78 & 70.56 & 64.99 \\
    &  & GPTQ & 6.147 & 45.82 & 72.26 & 74.10 & 76.24 & 70.15 & 43.80 & 80.20 & 71.90 & 66.81 \\
    &  & \bf{\boa} & \bf{5.189} & 47.44 & 74.62 & 77.31 & 78.03 & 73.19 & 45.40 & 80.20 & 73.09 & \bf{68.66} \\
    \midrule
    \multirowcell{11.8}{LLaMA-30B}
    & FP & Baseline & 4.101 & 52.90 & 78.96 & 82.69 & 82.63 & 75.47 & 48.20 & 82.26 & 75.77 & 72.36 \\
    \cmidrule{2-13}
    & \multirowcell{3}{INT2} 
    & RTN & 7.1e3 & 25.60 & 26.52 & 40.67 & 27.20 & 0.11 & 24.60 & 50.92 & 49.64 & 30.66 \\
    &  & GPTQ & 9.770 & 35.41 & 60.14 & 64.77 & 61.54 & 50.12 & 38.20 & 70.35 & 66.54 & 55.88 \\
    &  & \bf{\boa} & \bf{6.669} & 37.71 & 64.73 & 73.00 & 64.77 & 61.28 & 39.00 & 73.34 & 67.40 & \bf{60.15} \\
    \cmidrule{2-13}
    & \multirowcell{3}{INT3} 
    & RTN & 81.76 & 30.12 & 58.21 & 54.10 & 42.94 & 6.71 & 35.00 & 69.31 & 58.25 & 44.33 \\
    &  & GPTQ & 7.064 & 48.12 & 71.46 & 74.89 & 76.73 & 66.39 & 43.20 & 78.35 & 70.80 & 66.24 \\
    &  & \bf{\boa} & \bf{4.575} & 49.57 & 74.75 & 81.87 & 79.21 & 73.77 & 46.20 & 79.38 & 74.19 & \bf{69.87} \\
    \cmidrule{2-13}
    & \multirowcell{3}{INT4} 
    & RTN & 5.802 & 49.40 & 74.16 & 78.07 & 79.48 & 66.67 & 44.00 & 79.87 & 73.16 & 68.10 \\
    &  & GPTQ & 5.188 & 51.45 & 75.67 & 80.18 & 79.69 & 70.38 & 45.40 & 79.43 & 73.64 & 69.48 \\
    &  & \bf{\boa} & \bf{4.218} & 53.16 & 78.41 & 82.72 & 81.95 & 75.96 & 49.40 & 81.88 & 75.06 & \bf{72.32} \\
    \bottomrule
    \end{tabular}
    \end{threeparttable}

    \label{tab:weight_only_quant_without_transform_llama}
    
\end{table*}

\begin{table*}[!ht]
    \renewcommand{\arraystretch}{1.0}
    \fontsize{7.8pt}{9.36pt}\selectfont
    \centering
    \caption{Weight-only quantization performance on OPT models without transformation}
    \begin{threeparttable}
    \begin{tabular}{c c c c c c c c c c c c c}
    \toprule
    \multirowcell{2}{Model} & \multirowcell{2}{Precision} & \multirowcell{2}{Method} & \multirowcell{2}{Wiki2 PPL ($\downarrow$)} & \multicolumn{9}{c}{Zero-shot Accuracy ($\uparrow$)} \\
    \cline{5-13}
    & & & & Arc-c & Arc-e & BQ & HS & LAMB & OBQA & PIQA & WG & Average \\
    \toprule
    \multirowcell{8.5}{OPT-125M} 
    & FP & Baseline & 27.65 & 22.78 & 39.98 & 55.44 & 31.35 & 33.44 & 28.00 & 62.02 & 50.28 & 40.41 \\
    \cmidrule{2-13}
    & \multirowcell{3}{INT2} 
    & RTN & 1.0e4 & 23.89 & 27.31 & 37.83 & 25.96 & 0.00 & 27.00 & 50.49 & 51.14 & 30.45 \\
    &  & GPTQ & 411.3 & 22.61 & 31.52 & 38.47 & 27.16 & 0.18 & 27.20 & 52.18 & 51.70 & 31.38 \\
    &  & \bf{\boa} & \bf{85.63} & 21.42 & 32.15 & 48.20 & 28.50 & 6.86 & 26.60 & 56.47 & 49.17 & \bf{33.67} \\
    \cmidrule{2-13}
    & \multirowcell{3}{INT3} 
    & RTN & 233.9 & 22.01 & 33.33 & 52.94 & 28.52 & 2.81 & 28.40 & 59.19 & 51.14 & 34.79 \\
    &  & GPTQ & 50.75 & 21.67 & 33.84 & 58.38 & 29.41 & 16.96 & 25.40 & 58.98 & 51.14 & 36.97 \\
    &  & \bf{\boa} & \bf{31.95} & 22.95 & 36.83 & 59.66 & 30.67 & 27.10 & 28.80 & 60.34 & 50.83 & \bf{39.65} \\
    \midrule
    \multirowcell{8.5}{OPT-350M} 
    & FP & Baseline & 22.00 & 23.89 & 40.36 & 57.68 & 36.67 & 40.47 & 28.20 & 64.74 & 52.33 & 43.04 \\
    \cmidrule{2-13}
    & \multirowcell{3}{INT2} 
    & RTN & 4.6e3 & 24.74 & 27.65 & 37.83 & 26.87 & 0.00 & 23.40 & 52.83 & 52.09 & 30.68 \\
    &  & GPTQ & 61.93 & 22.53 & 33.46 & 57.28 & 29.66 & 10.70 & 23.80 & 56.75 & 51.22 & 35.68 \\
    &  & \bf{\boa} & \bf{46.53} & 22.01 & 35.90 & 62.32 & 30.43 & 11.85 & 26.60 & 58.98 & 51.62 & \bf{37.46} \\
    \cmidrule{2-13}
    & \multirowcell{3}{INT3} 
    & RTN & 35.91 & 23.38 & 36.99 & 52.14 & 34.00 & 30.37 & 27.00 & 62.02 & 51.62 & 39.69 \\
    &  & GPTQ & 25.25 & 22.95 & 38.47 & 60.58 & 35.25 & 41.38 & 27.60 & 62.46 & 51.22 & 42.49 \\
    &  & \bf{\boa} & \bf{23.96} & 22.87 & 38.76 & 61.22 & 35.30 & 38.08 & 28.40 & 62.57 & 53.28 & \bf{42.56} \\
    \midrule
    \multirowcell{8.5}{OPT-1.3B} 
    & FP & Baseline & 14.62 & 29.52 & 50.97 & 57.83 & 53.70 & 55.19 & 33.40 & 72.47 & 59.51 & 51.57 \\
    \cmidrule{2-13}
    & \multirowcell{3}{INT2} 
    & RTN & 1.6e4 & 24.23 & 24.75 & 51.19 & 26.51 & 0.00 & 29.00 & 52.29 & 50.04 & 32.25 \\
    &  & GPTQ & 177.4 & 22.78 & 31.78 & 57.06 & 28.13 & 6.37 & 24.80 & 53.75 & 48.70 & 34.17 \\
    &  & \bf{\boa} & \bf{31.65} & 22.70 & 37.67 & 62.17 & 34.65 & 18.02 & 26.80 & 61.48 & 51.78 & \bf{39.41} \\
    \cmidrule{2-13}
    & \multirowcell{3}{INT3} 
    & RTN & 754.8 & 25.34 & 34.72 & 52.17 & 36.31 & 10.25 & 27.20 & 60.07 & 51.78 & 37.23 \\
    &  & GPTQ & 19.35 & 27.13 & 43.43 & 54.25 & 47.23 & 36.30 & 29.40 & 68.01 & 54.30 & 45.01 \\
    &  & \bf{\boa} & \bf{15.77} & 27.39 & 48.74 & 58.99 & 50.09 & 51.50 & 30.60 & 69.37 & 57.30 & \bf{49.25} \\
    \midrule
    \multirowcell{8.5}{OPT-2.7B} 
    & FP & Baseline & 12.47 & 31.31 & 54.38 & 60.40 & 60.62 & 59.78 & 35.20 & 74.81 & 61.01 & 54.69 \\
    \cmidrule{2-13}
    & \multirowcell{3}{INT2} 
    & RTN & 2.7e5 & 26.11 & 26.73 & 62.17 & 26.77 & 0.00 & 29.20 & 50.87 & 51.07 & 34.12 \\
    &  & GPTQ & 135.5 & 24.66 & 33.71 & 60.28 & 33.62 & 17.62 & 27.00 & 57.73 & 50.59 & 38.15 \\
    &  & \bf{\boa} & \bf{22.30} & 25.77 & 41.62 & 62.26 & 42.00 & 33.76 & 28.80 & 62.08 & 52.96 & \bf{43.66} \\
    \cmidrule{2-13}
    & \multirowcell{3}{INT3} 
    & RTN & 170.5 & 25.26 & 38.34 & 56.51 & 36.64 & 14.61 & 29.80 & 66.27 & 53.04 & 40.06 \\
    &  & GPTQ & 15.64 & 29.27 & 49.07 & 45.90 & 51.79 & 42.92 & 33.00 & 69.80 & 57.54 & 47.41 \\
    &  & \bf{\boa} & \bf{13.63} & 29.10 & 52.90 & 65.29 & 55.86 & 56.94 & 34.00 & 72.03 & 60.06 & \bf{53.27} \\
    \midrule
    \multirowcell{8.5}{OPT-6.7B} 
    & FP & Baseline & 10.86 & 34.73 & 60.02 & 66.06 & 67.16 & 65.50 & 37.40 & 76.50 & 65.43 & 59.10 \\
    \cmidrule{2-13}
    & \multirowcell{3}{INT2} 
    & RTN & 2.9e4 & 25.00 & 25.38 & 37.83 & 26.66 & 0.00 & 29.20 & 50.00 & 51.46 & 30.69 \\
    &  & GPTQ & 95.95 & 23.21 & 33.75 & 61.77 & 32.09 & 16.25 & 25.80 & 57.02 & 52.01 & 37.74 \\
    &  & \bf{\boa} & \bf{20.57} & 28.24 & 47.31 & 61.80 & 47.00 & 34.09 & 33.20 & 66.87 & 57.54 & \bf{47.01} \\
    \cmidrule{2-13}
    & \multirowcell{3}{INT3} 
    & RTN & 53.76 & 27.73 & 40.87 & 50.95 & 48.67 & 28.32 & 31.20 & 68.17 & 53.12 & 43.63 \\
    &  & GPTQ & 12.05 & 32.51 & 56.40 & 54.53 & 62.49 & 55.99 & 36.60 & 74.21 & 60.85 & 54.20 \\
    &  & \bf{\boa} & \bf{11.25} & 32.76 & 57.32 & 66.85 & 64.53 & 63.64 & 37.60 & 75.57 & 64.09 & \bf{57.80} \\
    \midrule
    \multirowcell{8.5}{OPT-13B} 
    & FP & Baseline & 10.13 & 35.75 & 61.87 & 65.84 & 69.87 & 65.98 & 39.00 & 76.88 & 65.11 & 60.04 \\
    \cmidrule{2-13}
    & \multirowcell{3}{INT2} 
    & RTN & 4.7e4 & 26.79 & 26.64 & 40.40 & 26.25 & 0.00 & 27.40 & 49.78 & 50.04 & 30.91 \\
    &  & GPTQ & 49.40 & 26.28 & 37.04 & 62.14 & 41.60 & 29.17 & 27.60 & 61.04 & 52.17 & 42.13 \\
    &  & \bf{\boa} & \bf{15.34} & 29.27 & 48.95 & 62.78 & 53.63 & 44.71 & 32.60 & 68.50 & 58.64 & \bf{49.89} \\
    \cmidrule{2-13}
    & \multirowcell{3}{INT3}
    & RTN & 40.46 & 27.47 & 38.72 & 62.54 & 51.21 & 29.44 & 26.80 & 63.38 & 49.88 & 43.68 \\
    &  & GPTQ & 11.11 & 34.56 & 59.51 & 62.45 & 64.81 & 58.57 & 36.00 & 74.92 & 63.54 & 56.80 \\
    &  & \bf{\boa} & \bf{10.39} & 34.30 & 59.93 & 69.94 & 66.95 & 66.61 & 37.60 & 75.95 & 66.93 & \bf{59.78} \\
    \midrule
    \multirowcell{8.5}{OPT-30B} 
    & FP & Baseline & 9.557 & 38.05 & 65.36 & 70.46 & 72.27 & 67.85 & 40.20 & 78.07 & 68.43 & 62.59 \\
    \cmidrule{2-13}
    & \multirowcell{3}{INT2} 
    & RTN & 4.2e4 & 26.79 & 25.55 & 37.83 & 25.88 & 0.00 & 28.60 & 50.60 & 51.22 & 30.81 \\
    &   & GPTQ & 21.19 & 27.99 & 41.96 & 62.45 & 46.90 & 39.48 & 30.40 & 65.72 & 54.22 & 46.14 \\
    &   & \bf{\boa} & \bf{11.15} & 30.55 & 55.81 & 62.81 & 60.19 & 60.39 & 34.80 & 72.47 & 62.67 & \bf{54.96} \\
    \cmidrule{2-13}
    & \multirowcell{3}{INT3}
    & RTN & 86.63 & 24.32 & 35.06 & 57.98 & 38.07 & 12.17 & 26.00 & 59.90 & 52.09 & 38.20 \\
    &   & GPTQ & 10.14 & 34.90 & 61.70 & 66.97 & 69.34 & 63.70 & 40.40 & 77.26 & 67.01 & 60.16 \\
    &   & \bf{\boa} & \bf{9.744} & 36.26 & 62.63 & 70.28 & 70.52 & 68.86 & 40.20 & 77.75 & 67.17 & \bf{61.71} \\
    \bottomrule
    \end{tabular}
    \end{threeparttable}
    \label{tab:weight_only_quant_without_transform_opt}
\end{table*}

\newpage

\subsection{Comparison with Transformation-based Methods}  \label{appendix:weight_only_quant_with_transformation}

We compare the weight-only quantization performances of the proposed \boa \ and existing transformation-based methods.
For conventional methods, we ran the official codes provided by the authors and reported the obtained results.
For OmniQuant~\citep{shao2023omniquant} and AffineQuant~\citep{ma2024affinequant}, we activated both learnable equivalent transformation (LET) and learnable weight clipping (LWC) options.
For AffineQuant, we did not activate the \texttt{use-ln-matrix} option because this adds extra affine transformation between the normalization layer and linear layers, which incurs additional processing time during the inference.
QuaRot-RTN and QuaRot-GPTQ mean that RTN and GPTQ have been applied for the weight quantization after transforming models via QuaRot~\citep{ashkboos2024quarot}.
For DuQuant, we report the results obtained with and without activating the LWC option as in the original paper~\citep{lin2024duquant}.
When measuring the performance of the proposed \boa, we also transformed the model for fair comparison.
We applied QuaRot for the transformation because QuaRot does not require training and incurs no extra costs during the actual inference~\citep{ashkboos2024quarot}.

In this appendix, we provide the results for the INT3 weight quantization of LLaMA2 and LLaMA3 models (see \cref{tab:weight_only_quant_with_transform_int3}) and the results on LLaMA1 models (see \cref{tab:weight_only_quant_with_transform_llama}).

\begin{table*}[ht]
    \renewcommand{\arraystretch}{1.0}
    \scriptsize
    \centering
    \caption{INT3 weight-only quantization performance on transformed LLaMA2 and LLaMA3 models}
    \begin{threeparttable}
    \begin{tabular}{c c c c c c c c c c c c c}
    \toprule
    \multirowcell{2}{Model} & \multirowcell{2}{Precision} & \multirowcell{2}{Method} & \multirowcell{2}{Wiki2 PPL ($\downarrow$)} & \multicolumn{9}{c}{Zero-shot Accuracy ($\uparrow$)} \\
    \cline{5-13}
    & & & & Arc-c & Arc-e & BQ & HS & LAMB & OBQA & PIQA & WG & Average \\
    \toprule
    \multirowcell{9.5}{LLaMA2-7B}
    & FP16 & Baseline & 5.473 & 45.90 & 74.66 & 77.92 & 75.94 & 70.86 & 44.00 & 78.89 & 68.90 & 67.13 \\
    \cmidrule{2-13}
    & \multirowcell{7}{INT3}
    & OmniQuant & 6.640 & 39.59 & 65.66 & 69.82 & 70.09 & 57.58 & 38.40 & 76.01 & 64.88 & 60.25 \\
    & & AffineQuant & 6.468 & 40.36 & 68.01 & 70.86 & 70.32 & 61.87 & 38.60 & 76.12 & 65.27 & 61.43 \\
    & & QuaRot-RTN & 129.2 & 23.46 & 30.60 & 37.83 & 29.30 & 4.75 & 24.40 & 53.10 & 50.20 & 31.71 \\
    & & QuaRot-GPTQ & 6.122 & 40.96 & 70.16 & 73.30 & 71.44 & 67.63 & 41.40 & 77.42 & 67.48 & 63.72 \\
    & & DuQuant & 6.831 & 41.13 & 67.59 & 69.36 & 70.52 & 62.25 & 41.40 & 76.01 & 65.35 & 61.70 \\
    & & DuQuant + LWC & 6.226 & 40.70 & 69.15 & 74.77 & 70.97 & 64.06 & 42.40 & 77.04 & 66.85 & 63.24 \\
    & & \bf{\boa}$^{\dagger}$ & \bf{5.874} & 42.06 & 71.17 & 73.73 & 72.29 & 69.85 & 41.60 & 77.37 & 67.48 & \bf{64.44} \\
    \midrule
    \multirowcell{9.5}{LLaMA2-13B}
    & FP16 & Baseline & 4.885 & 49.06 & 77.65 & 80.49 & 79.38 & 73.41 & 45.80 & 80.69 & 72.22 & 69.84 \\
    \cmidrule{2-13}
    & \multirowcell{7}{INT3}
    & OmniQuant & 5.593 & 44.80 & 71.72 & 75.29 & 75.63 & 64.94 & 43.20 & 78.67 & 69.30 & 65.44 \\
    & & AffineQuant & 5.526 & 45.73 & 73.74 & 77.34 & 74.79 & 65.56 & 43.00 & 77.15 & 67.01 & 65.54 \\
    & & QuaRot-RTN & 48.06 & 22.01 & 34.72 & 62.14 & 33.44 & 7.80 & 26.60 & 57.67 & 50.51 & 36.86 \\
    & & QuaRot-GPTQ & 5.382 & 47.01 & 75.42 & 78.84 & 75.70 & 72.14 & 44.60 & 78.56 & 70.01 & 67.79 \\
    & & DuQuant & 5.749 & 44.88 & 72.10 & 79.91 & 75.37 & 68.23 & 43.00 & 77.80 & 70.72 & 66.50 \\
    & & DuQuant + LWC & 5.414 & 46.67 & 74.62 & 77.92 & 75.93 & 68.63 & 43.80 & 79.98 & 68.75 & 67.04 \\
    & & \bf{\boa}$^{\dagger}$ & \bf{5.202} & 48.29 & 76.39 & 79.45 & 76.23 & 73.19 & 45.20 & 78.67 & 70.96 & \bf{68.55} \\
    \midrule
    \multirowcell{9.5}{LLaMA3-8B}
    & FP16 & Baseline & 6.137 & 53.67 & 77.61 & 81.19 & 79.15 & 72.23 & 45.00 & 81.01 & 73.24 & 70.39 \\
    \cmidrule{2-13}
    & \multirowcell{7}{INT3}
    & OmniQuant & 11.75 & 37.37 & 60.94 & 65.57 & 66.81 & 33.02 & 34.80 & 71.49 & 59.59 & 53.70 \\
    & & AffineQuant & 11.63 & 37.97 & 63.64 & 64.92 & 67.42 & 33.95 & 36.20 & 71.98 & 60.30 & 54.55 \\
    & & QuaRot-RTN & 38.64 & 23.72 & 38.22 & 51.28 & 47.53 & 23.63 & 33.60 & 62.35 & 62.04 & 42.80 \\
    & & QuaRot-GPTQ & 7.490 & 47.87 & 74.49 & 78.44 & 74.67 & 67.43 & 40.00 & 78.07 & 72.38 & 66.67 \\
    & & DuQuant & 11.35 & 35.84 & 53.37 & 64.56 & 65.56 & 49.52 & 36.20 & 72.25 & 65.90 & 55.40 \\
    & & DuQuant + LWC & 10.78 & 31.66 & 46.59 & 64.65 & 65.25 & 41.29 & 38.00 & 70.13 & 61.80 & 52.42 \\
    & & \bf{\boa}$^{\dagger}$ & \bf{7.145} & 49.06 & 77.40 & 80.49 & 74.54 & 69.20 & 43.00 & 78.51 & 72.53 & \bf{68.09} \\
    \midrule
    \multirowcell{9.5}{LLaMA3.2-1B}
    & FP16 & Baseline & 13.15 & 38.14 & 63.26 & 69.51 & 60.78 & 54.38 & 34.60 & 74.37 & 59.51 & 56.82 \\
    \cmidrule{2-13}
    & \multirowcell{7}{INT3}
    & OmniQuant & 26.61 & 31.31 & 51.39 & 63.67 & 49.18 & 18.83 & 29.60 & 67.57 & 55.49 & 45.88 \\
    & & AffineQuant & 26.43 & 32.00 & 52.48 & 64.31 & 48.71 & 20.67 & 26.20 & 67.57 & 52.88 & 45.60 \\
    & & QuaRot-RTN & 98.24 & 24.06 & 36.03 & 61.96 & 35.31 & 7.68 & 28.60 & 59.14 & 51.62 & 38.05 \\
    & & QuaRot-GPTQ & 16.56 & 32.42 & 57.95 & 64.10 & 52.48 & 43.79 & 32.20 & 68.99 & 57.14 & 51.13 \\
    & & DuQuant & 2.1e4 & 25.26 & 25.67 & 41.35 & 26.68 & 0.00 & 27.80 & 49.51 & 48.78 & 30.63 \\
    & & DuQuant + LWC & 2.7e4 & 25.68 & 26.73 & 40.73 & 25.34 & 0.00 & 29.00 & 49.40 & 52.09 & 31.12 \\
    & & \bf{\boa}$^{\dagger}$ & \bf{15.73} & 32.42 & 57.95 & 66.24 & 53.86 & 48.74 & 33.20 & 70.18 & 57.06 & \bf{52.46} \\
    \midrule
    \multirowcell{9.5}{LLaMA3.2-3B}
    & FP16 & Baseline & 11.04 & 46.16 & 67.80 & 78.62 & 70.44 & 62.15 & 36.00 & 75.52 & 67.40 & 63.01 \\
    \cmidrule{2-13}
    & \multirowcell{7}{INT3}
    & OmniQuant & 16.97 & 36.01 & 57.58 & 71.47 & 59.57 & 39.17 & 34.00 & 69.31 & 60.22 & 53.42 \\
    & & AffineQuant & 16.79 & 36.43 & 56.57 & 72.97 & 59.19 & 40.07 & 35.20 & 70.13 & 60.06 & 53.83 \\
    & & QuaRot-RTN & 89.54 & 22.01 & 32.79 & 59.08 & 36.18 & 4.82 & 25.00 & 54.35 & 53.67 & 35.99 \\
    & & QuaRot-GPTQ & 13.58 & 38.31 & 58.96 & 75.72 & 64.59 & 54.59 & 35.80 & 70.18 & 64.96 & 57.89 \\
    & & DuQuant & 18.92 & 30.55 & 48.32 & 71.71 & 60.34 & 40.54 & 32.40 & 66.54 & 60.38 & 51.35 \\
    & & DuQuant + LWC & 15.18 & 37.20 & 58.33 & 72.02 & 61.22 & 44.58 & 33.40 & 70.02 & 59.98 & 54.59 \\
    & & \bf{\boa}$^{\dagger}$ & \bf{12.97} & 42.49 & 65.74 & 78.90 & 65.59 & 58.40 & 34.80 & 73.07 & 63.46 & \bf{60.31} \\
    \bottomrule
    \end{tabular}
    \begin{tablenotes}
        \item[$\dagger$] \boa \ has been applied after transforming the model via QuaRot.
        \item[*] The LET option has been deactivated because this option does not support models exploiting GQA.
    \end{tablenotes}
    \end{threeparttable}
    \label{tab:weight_only_quant_with_transform_int3}

    \vspace{-.3cm}
    
\end{table*}

\begin{table*}[!ht]
    \renewcommand{\arraystretch}{1.0}
    \scriptsize
    \centering
    \caption{Weight-only quantization performance on transformed LLaMA models}
    \begin{threeparttable}
    \begin{tabular}{c c c c c c c c c c c c c}
    \toprule
    \multirowcell{2}{Model} & \multirowcell{2}{Precision} & \multirowcell{2}{Method} & \multirowcell{2}{Wiki2 PPL ($\downarrow$)} & \multicolumn{9}{c}{Zero-shot Accuracy ($\uparrow$)} \\
    \cline{5-13}
    & & & & Arc-c & Arc-e & BQ & HS & LAMB & OBQA & PIQA & WG & Average \\
    \toprule
    \multirowcell{16}{LLaMA-7B}
    & FP & Baseline & 5.677 & 44.71 & 72.94 & 75.05 & 76.21 & 70.66 & 44.40 & 79.16 & 70.01 & 66.64 \\
    \cmidrule{2-13}
    & \multirowcell{7}{INT2} 
    & OmniQuant & 15.74 & 26.28 & 45.71 & 61.93 & 42.66 & 17.43 & 29.00 & 62.40 & 52.17 & 42.20 \\
    &  & AffineQuant & 11.93 & 27.22 & 49.33 & 62.51 & 49.26 & 23.57 & 32.20 & 62.89 & 54.06 & 45.13 \\
    &  & QuaRot-RTN & 1.0e4 & 27.90 & 26.39 & 37.83 & 26.03 & 0.00 & 21.60 & 49.24 & 49.49 & 29.81 \\
    &  & QuaRot-GPTQ & 11.53 & 26.45 & 45.75 & 62.48 & 48.23 & 37.37 & 32.80 & 65.67 & 57.46 & 47.03 \\
    &  & DuQuant & 9.4e3 & 28.33 & 26.89 & 46.42 & 26.24 & 0.00 & 21.20 & 50.71 & 51.62 & 31.43 \\
    &  & DuQuant + LWC & 12.39 & 27.73 & 49.37 & 63.03 & 49.88 & 23.65 & 31.80 & 64.96 & 56.75 & 45.90 \\
    &  & \bf{\boa}$^{\dagger}$ & \bf{9.812} & 30.63 & 54.88 & 63.55 & 52.20 & 46.50 & 33.00 & 68.72 & 61.01 & \bf{51.31} \\
    \cmidrule{2-13}
    & \multirowcell{7}{INT3} 
    & OmniQuant & 6.463 & 38.74 & 66.96 & 72.35 & 69.84 & 62.79 & 40.00 & 77.20 & 66.69 & 61.82 \\
    &  & AffineQuant & 6.392 & 39.68 & 66.46 & 72.87 & 70.52 & 64.99 & 39.40 & 77.42 & 65.04 & 62.05 \\
    &  & QuaRot-RTN & 24.03 & 29.86 & 53.11 & 61.13 & 50.30 & 24.53 & 31.40 & 69.37 & 57.30 & 47.13 \\
    &  & QuaRot-GPTQ & 6.166 & 43.09 & 70.37 & 72.02 & 72.18 & 65.34 & 43.20 & 78.29 & 68.19 & 64.09 \\
    &  & DuQuant & 6.976 & 37.37 & 64.23 & 71.13 & 70.81 & 60.45 & 39.20 & 76.39 & 65.98 & 60.70 \\
    &  & DuQuant + LWC & 6.328 & 38.74 & 66.20 & 73.73 & 71.24 & 63.78 & 42.40 & 77.31 & 68.51 & 62.74 \\
    &  & \bf{\boa}$^{\dagger}$ & \bf{6.091} & 41.04 & 68.18 & 71.77 & 72.26 & 69.32 & 40.40 & 78.18 & 68.67 & \bf{63.73} \\
    \midrule
    \multirowcell{16}{LLaMA-13B}
    & FP & Baseline & 5.090 & 47.70 & 74.75 & 77.95 & 79.07 & 73.66 & 44.80 & 80.14 & 72.77 & 68.86 \\
    \cmidrule{2-13}
    & \multirowcell{7}{INT2} 
    & OmniQuant & 13.29 & 29.69 & 53.79 & 62.29 & 55.31 & 19.52 & 31.60 & 70.29 & 57.14 & 47.45 \\
    &  & AffineQuant & 10.65 & 28.75 & 53.54 & 65.57 & 53.03 & 30.73 & 34.00 & 68.12 & 59.35 & 49.14 \\
    &  & QuaRot-RTN & 4.1e3 & 28.16 & 26.47 & 37.83 & 25.72 & 0.00 & 22.60 & 51.25 & 48.46 & 30.06 \\
    &  & QuaRot-GPTQ & 9.331 & 31.06 & 58.59 & 63.79 & 54.96 & 45.54 & 35.20 & 69.97 & 63.30 & 52.80 \\
    &  & DuQuant & 6.2e3 & 25.94 & 25.93 & 39.66 & 26.70 & 0.00 & 23.60 & 50.87 & 51.62 & 30.54 \\
    &  & DuQuant + LWC & 8.770 & 31.74 & 58.38 & 66.27 & 60.98 & 44.46 & 37.00 & 72.20 & 62.19 & 54.15 \\
    &  & \bf{\boa}$^{\dagger}$ & \bf{8.341} & 33.19 & 63.64 & 71.44 & 60.75 & 56.30 & 37.00 & 71.93 & 65.90 & \bf{57.52} \\
    \cmidrule{2-13}
    & \multirowcell{7}{INT3} 
    & OmniQuant & 5.671 & 45.05 & 71.13 & 75.44 & 75.35 & 66.82 & 44.40 & 78.56 & 69.46 & 65.78 \\
    &  & AffineQuant & 5.615 & 43.00 & 70.79 & 74.83 & 74.99 & 68.95 & 43.40 & 79.76 & 69.38 & 65.64 \\
    &  & QuaRot-RTN & 7.082 & 37.37 & 60.14 & 73.18 & 67.93 & 56.42 & 35.60 & 76.22 & 66.14 & 59.13 \\
    &  & QuaRot-GPTQ & 5.471 & 44.37 & 71.63 & 76.94 & 75.59 & 72.45 & 43.00 & 78.02 & 71.51 & 66.69 \\
    &  & DuQuant & 5.923 & 43.94 & 70.12 & 74.40 & 75.76 & 68.33 & 42.00 & 78.56 & 72.06 & 65.65 \\
    &  & DuQuant + LWC & 5.554 & 45.31 & 71.97 & 72.78 & 75.85 & 69.81 & 45.00 & 79.49 & 70.64 & 66.36 \\
    &  & \bf{\boa}$^{\dagger}$ & \bf{5.411} & 45.56 & 72.81 & 75.60 & 75.95 & 72.71 & 45.20 & 79.27 & 71.03 & \bf{67.27} \\
    \midrule
    \multirowcell{16}{LLaMA-30B}
    & FP & Baseline & 4.101 & 52.90 & 78.96 & 82.69 & 82.63 & 75.47 & 48.20 & 82.26 & 75.77 & 72.36 \\
    \cmidrule{2-13}
    & \multirowcell{7}{INT2} 
    & OmniQuant & 8.598 & 33.45 & 57.37 & 62.84 & 59.21 & 39.36 & 37.20 & 70.02 & 60.14 & 52.45 \\
    &  & AffineQuant & 7.267 & 38.14 & 65.07 & 73.33 & 67.61 & 53.75 & 38.60 & 74.21 & 64.96 & 59.46 \\
    &  & QuaRot-RTN & 3.8e3 & 25.77 & 26.52 & 37.83 & 25.80 & 0.01 & 23.80 & 51.69 & 47.83 & 29.91 \\
    &  & QuaRot-GPTQ & 7.283 & 39.08 & 68.01 & 65.44 & 65.54 & 58.15 & 38.40 & 73.39 & 67.56 & 59.45 \\
    &  & DuQuant & 5.7e3 & 27.13 & 26.98 & 38.75 & 26.76 & 0.00 & 26.20 & 49.89 & 49.17 & 30.61 \\
    &  & DuQuant + LWC & 7.706 & 38.48 & 64.60 & 70.00 & 67.56 & 48.08 & 36.40 & 72.96 & 65.11 & 57.90 \\
    &  & \bf{\boa}$^{\dagger}$ & \bf{6.525} & 41.47 & 68.01 & 64.65 & 67.63 & 66.31 & 41.00 & 74.21 & 70.32 & \bf{61.70} \\
    \cmidrule{2-13}
    & \multirowcell{7}{INT3} 
    & OmniQuant & 4.766 & 50.09 & 76.18 & 79.91 & 79.58 & 70.79 & 45.20 & 79.92 & 74.43 & 69.51 \\
    &  & AffineQuant & 4.729 & 50.60 & 76.81 & 79.85 & 79.60 & 72.83 & 44.60 & 80.52 & 74.74 & 69.94 \\
    &  & QuaRot-RTN & 6.355 & 37.80 & 65.32 & 67.98 & 64.79 & 47.46 & 38.80 & 72.80 & 67.17 & 57.77 \\
    &  & QuaRot-GPTQ & 4.759 & 49.23 & 77.19 & 81.22 & 79.97 & 75.98 & 45.20 & 80.41 & 75.61 & 70.60 \\
    &  & DuQuant & 5.066 & 49.49 & 74.71 & 79.97 & 79.87 & 72.78 & 46.40 & 79.54 & 74.51 & 69.66 \\
    &  & DuQuant + LWC & 4.634 & 49.32 & 75.17 & 81.35 & 80.26 & 72.29 & 46.20 & 81.01 & 73.72 & 69.92 \\
    &  & \bf{\boa}$^{\dagger}$ & \bf{4.602} & 52.73 & 77.23 & 80.83 & 80.04 & 75.91 & 45.60 & 80.79 & 76.16 & \bf{71.16} \\
    \bottomrule
    \end{tabular}
    \begin{tablenotes}
        \item[$\dagger$] \boa \ has been applied after transforming the model via QuaRot.
    \end{tablenotes}
    \end{threeparttable}
    \label{tab:weight_only_quant_with_transform_llama}
    \vspace{5cm}
\end{table*}

\newpage

\subsection{Weight-Activation Quantization Results}  \label{appendix:weight_act_quant}

We evaluate the weight-activation quantization performances of the proposed \boa.
As in prior works~\cite{shao2023omniquant, ma2024affinequant, lin2024duquant, ashkboos2024quarot, liu2024spinquant}, we apply per-token nearest-rounding quantization for input activations and KV caches.
We use the notation `WxAyKVz' to denote the x-bit weight quantization, y-bit input activation quantization, and z-bit KV cache quantization.
When measuring the performance of RTN, GPTQ, and the proposed \boa, we use SpinQuant for the model transformation as SpinQuant outperforms QuaRot by optimizing the transformation for the activation quantization~\cite{liu2024spinquant}.
When optimizing the rotation matrix in SpinQuant, we quantize both weights and activations for SpinQuant-RTN and quantize only activations for SpinQuant-GPTQ and the proposed \boa, as in the original paper~\citep{liu2024spinquant}.
In this appendix, we provide the results on LLaMA3 models for various quantization configurations.

\begin{table*}[!ht]
    \renewcommand{\arraystretch}{1.0}
    \fontsize{6.25pt}{7.5pt}\selectfont
    \centering
    \vspace{-.25cm}
    \caption{Weight-activation quantization performance on the transformed LLaMA3-8B}
    \begin{threeparttable}
    \begin{tabular}{c c c c c c c c c c c c c}
    \toprule
    \multirowcell{2}{Precision} & \multirowcell{2}{Method} & \multirowcell{2}{Wiki2 PPL ($\downarrow$)} & \multicolumn{9}{c}{Zero-shot Accuracy ($\uparrow$)} \\
    \cline{4-12}
    & & & Arc-c & Arc-e & BQ & HS & LAMB & OBQA & PIQA & WG & Average \\
    \toprule
    FP16 & Baseline & 6.137 & 53.67 & 77.61 & 81.19 & 79.15 & 72.23 & 45.00 & 81.01 & 73.24 & 70.39 \\
    \midrule
    \multirowcell{7}{W2A4KV16} 
    & OmniQuant & 2.9e5 & 25.94 & 26.22 & 38.50 & 24.86 & 0.00 & 28.00 & 50.44 & 49.33 & 30.41 \\
    & AffineQuant & 3.2e5 & 26.62 & 26.30 & 38.13 & 25.62 & 0.00 & 29.60 & 49.51 & 52.57 & 31.04 \\
    & SpinQuant-RTN & 28.38 & 25.77 & 35.52 & 62.11 & 37.55 & 18.89 & 24.20 & 56.42 & 51.85 & 39.04 \\
    & SpinQuant-GPTQ & 26.35 & 23.38 & 38.30 & 60.43 & 37.34 & 16.44 & 26.20 & 58.71 & 53.51 & 39.29 \\
    & DuQuant & 4.5e5 & 25.77 & 25.97 & 43.09 & 26.09 & 0.00 & 27.00 & 51.85 & 50.83 & 31.33 \\
    & DuQuant + LWC & 4.4e4 & 26.28 & 27.10 & 38.38 & 25.81 & 0.00 & 28.00 & 50.49 & 50.12 & 30.77 \\
    & \bf{\boa}$^{\dagger}$ & \bf{17.31} & 27.82 & 47.52 & 63.27 & 44.54 & 27.27 & 29.40 & 61.86 & 54.54 & \bf{44.53} \\
    \midrule
    \multirowcell{7}{W3A3KV16} 
    & OmniQuant & 2.7e4 & 26.79 & 26.05 & 38.62 & 25.43 & 0.00 & 26.20 & 50.16 & 50.36 & 30.45 \\
    & AffineQuant & 3.0e4 & 26.45 & 25.25 & 38.62 & 25.20 & 0.00 & 30.00 & 50.76 & 49.41 & 30.71 \\
    & SpinQuant-RTN & 21.27 & 27.05 & 42.93 & 62.75 & 45.02 & 27.07 & 29.00 & 62.02 & 52.25 & 43.51 \\
    & SpinQuant-GPTQ & 20.49 & 26.96 & 40.40 & 54.77 & 50.47 & 26.17 & 31.80 & 62.13 & 53.35 & 43.26 \\
    & DuQuant & 280.8 & 23.29 & 29.42 & 41.41 & 30.84 & 1.37 & 26.40 & 51.09 & 51.22 & 31.88 \\
    & DuQuant + LWC & 783.1 & 22.35 & 27.57 & 38.07 & 28.05 & 0.18 & 27.00 & 51.74 & 50.59 & 30.69 \\
    & \bf{\boa}$^{\dagger}$ & \bf{17.56} & 31.66 & 45.88 & 57.98 & 53.55 & 32.56 & 32.00 & 65.34 & 52.17 & \bf{46.39} \\
    \midrule
    \multirowcell{7}{W4A4KV16} 
    & OmniQuant & 142.4 & 22.44 & 35.02 & 41.65 & 33.69 & 2.46 & 27.60 & 54.90 & 51.14 & 33.61 \\
    & AffineQuant & 144.5 & 25.43 & 32.70 & 45.17 & 33.72 & 2.83 & 25.20 & 56.58 & 49.96 & 33.95 \\
    & SpinQuant-RTN & 8.229 & 45.14 & 72.10 & 75.44 & 74.07 & 59.53 & 40.80 & 76.44 & 67.88 & 63.93 \\
    & SpinQuant-GPTQ & 7.636 & 46.33 & 72.31 & 75.17 & 72.69 & 64.27 & 42.80 & 76.44 & 68.27 & 64.79 \\
    & DuQuant & 7.793 & 45.90 & 71.46 & 72.60 & 74.36 & 66.06 & 42.40 & 78.02 & 69.61 & 65.05 \\
    & DuQuant + LWC & 8.066 & 44.20 & 71.13 & 74.01 & 73.59 & 58.55 & 40.60 & 76.33 & 66.77 & 63.15 \\
    & \bf{\boa}$^{\dagger}$ & \bf{7.496} & 47.10 & 72.05 & 74.98 & 74.22 & 66.29 & 42.00 & 76.33 & 69.53 & \bf{65.31} \\
    \midrule
    \multirowcell{7}{W4A4KV4} 
    & OmniQuant & 188.2 & 22.78 & 32.32 & 43.39 & 31.69 & 2.01 & 25.40 & 54.79 & 50.20 & 32.82 \\
    & AffineQuant & 221.5 & 22.61 & 31.23 & 40.98 & 31.06 & 1.53 & 24.60 & 55.77 & 50.28 & 32.26 \\
    & SpinQuant-RTN & 8.503 & 42.06 & 69.40 & 72.78 & 72.69 & 60.50 & 38.40 & 74.21 & 65.19 & 61.90 \\
    & SpinQuant-GPTQ & 7.869 & 45.56 & 73.32 & 71.07 & 73.88 & 63.13 & 39.80 & 77.04 & 68.59 & 64.05 \\
    & DuQuant & 8.000 & 45.99 & 70.88 & 74.01 & 73.67 & 64.32 & 41.00 & 76.55 & 68.90 & 64.42 \\
    & DuQuant + LWC & 8.402 & 44.80 & 70.29 & 73.82 & 73.69 & 58.61 & 39.00 & 75.08 & 66.77 & 62.76 \\
    & \bf{\boa}$^{\dagger}$ & \bf{7.705} & 45.22 & 74.54 & 72.81 & 74.09 & 65.72 & 43.60 & 77.53 & 66.69 & \bf{65.03} \\
    \bottomrule
    \end{tabular}
    \begin{tablenotes}
        \item[$\dagger$] \boa \ has been applied after transforming the model via SpinQuant.
        \item[*] The LET option has been deactivated for OmniQuant and AffineQuant because this option does not support models exploiting GQA.
    \end{tablenotes}
    \end{threeparttable}
    \label{tab:weight_act_quant_llama3-8b}
    \vspace{-.25cm}
\end{table*}

\begin{table*}[!ht]
    \renewcommand{\arraystretch}{1.0}
    \fontsize{6.25pt}{7.5pt}\selectfont
    \centering
    \caption{Weight-activation quantization performance on the transformed LLaMA3.2-1B}
    \begin{threeparttable}
    \begin{tabular}{c c c c c c c c c c c c c}
    \toprule
    \multirowcell{2}{Precision} & \multirowcell{2}{Method} & \multirowcell{2}{Wiki2 PPL ($\downarrow$)} & \multicolumn{9}{c}{Zero-shot Accuracy ($\uparrow$)} \\
    \cline{4-12}
    & & & Arc-c & Arc-e & BQ & HS & LAMB & OBQA & PIQA & WG & Average \\
    \toprule
    FP16 & Baseline & 13.15 & 38.14 & 63.26 & 69.51 & 60.78 & 54.38 & 34.60 & 74.37 & 59.51 & 56.82 \\
    \midrule
    \multirowcell{7}{W2A4KV16} 
    & OmniQuant & 2.5e3 & 24.83 & 26.85 & 37.83 & 26.01 & 0.00 & 29.00 & 52.34 & 51.78 & 31.08 \\
    & AffineQuant & 6.1e3 & 25.00 & 26.73 & 37.89 & 25.64 & 0.00 & 26.80 & 52.18 & 50.36 & 30.58 \\
    & SpinQuant-RTN & 93.90 & 23.63 & 32.62 & 54.59 & 28.71 & 7.51 & 24.60 & 53.43 & 50.75 & 34.48 \\
    & SpinQuant-GPTQ & 104.4 & 21.59 & 32.28 & 50.83 & 30.11 & 7.55 & 26.20 & 53.75 & 49.96 & 34.03 \\
    & DuQuant & 1.0e5 & 26.28 & 25.04 & 46.61 & 26.39 & 0.00 & 29.40 & 49.13 & 47.51 & 31.30 \\
    & DuQuant + LWC & 1.0e4 & 26.88 & 24.54 & 38.04 & 25.78 & 0.00 & 28.00 & 50.98 & 50.43 & 30.58 \\
    & \bf{\boa}$^{\dagger}$ & \bf{59.95} & 23.98 & 37.04 & 54.68 & 32.02 & 9.33 & 27.00 & 56.04 & 52.41 & \bf{36.56} \\
    \midrule
    \multirowcell{7}{W3A3KV16} 
    & OmniQuant & 7.1e3 & 26.79 & 26.60 & 38.20 & 25.24 & 0.00 & 29.20 & 51.03 & 48.93 & 30.75 \\
    & AffineQuant & 5.6e3 & 27.39 & 26.77 & 38.38 & 25.59 & 0.00 & 29.40 & 49.89 & 49.09 & 30.81 \\
    & SpinQuant-RTN & 57.43 & 24.23 & 33.75 & 53.76 & 32.59 & 12.11 & 25.40 & 53.86 & 49.72 & 35.68 \\
    & SpinQuant-GPTQ & 57.52 & 23.38 & 36.07 & 48.93 & 35.13 & 10.89 & 27.40 & 55.82 & 47.12 & 35.59 \\
    & DuQuant & 2.9e4 & 25.09 & 26.14 & 40.00 & 25.94 & 0.01 & 28.40 & 50.49 & 50.75 & 30.85 \\
    & DuQuant + LWC & 1.2e4 & 25.17 & 25.21 & 39.33 & 25.65 & 0.00 & 27.20 & 51.85 & 51.22 & 30.70 \\
    & \bf{\boa}$^{\dagger}$ & \bf{48.47} & 23.81 & 38.97 & 53.21 & 35.55 & 13.90 & 26.80 & 54.84 & 52.49 & \bf{37.45} \\
    \midrule
    \multirowcell{7}{W4A4KV16} 
    & OmniQuant & 156.0 & 24.23 & 32.66 & 47.89 & 31.15 & 0.84 & 27.20 & 54.84 & 50.12 & 33.62 \\
    & AffineQuant & 155.8 & 23.63 & 35.14 & 47.86 & 30.61 & 0.97 & 28.80 & 54.52 & 48.30 & 33.73 \\
    & SpinQuant-RTN & 17.66 & 32.25 & 54.80 & 63.30 & 53.38 & 40.69 & 30.80 & 68.93 & 53.75 & 49.74 \\
    & SpinQuant-GPTQ & 16.68 & 33.79 & 55.56 & 64.77 & 55.08 & 41.35 & 33.40 & 68.28 & 54.85 & 50.89 \\
    & DuQuant & 2.3e4 & 26.37 & 26.47 & 40.28 & 26.40 & 0.01 & 30.00 & 48.59 & 48.38 & 30.81 \\
    & DuQuant + LWC & 1.9e4 & 26.19 & 26.81 & 40.31 & 25.48 & 0.00 & 26.00 & 50.76 & 47.12 & 30.33 \\
    & \bf{\boa}$^{\dagger}$ & \bf{16.25} & 33.62 & 58.12 & 65.72 & 55.06 & 44.28 & 31.80 & 69.80 & 55.64 & \bf{51.76} \\
    \midrule
    \multirowcell{7}{W4A4KV4} 
    & OmniQuant & 219.2 & 23.55 & 31.82 & 45.47 & 29.46 & 0.79 & 28.00 & 51.52 & 48.22 & 32.35 \\
    & AffineQuant & 222.0 & 23.46 & 32.74 & 48.53 & 29.17 & 0.64 & 30.20 & 52.88 & 52.64 & 33.78 \\
    & SpinQuant-RTN & 19.68 & 32.25 & 52.86 & 61.71 & 50.74 & 34.91 & 31.60 & 66.65 & 53.43 & 48.02 \\
    & SpinQuant-GPTQ & 18.31 & 30.72 & 55.18 & 61.90 & 52.91 & 39.13 & 31.80 & 67.30 & 51.93 & 48.86 \\
    & DuQuant & 2.0e4 & 25.60 & 25.97 & 39.91 & 26.03 & 0.00 & 27.80 & 49.02 & 50.36 & 30.59 \\
    & DuQuant + LWC & 1.7e4 & 27.05 & 26.01 & 38.96 & 26.07 & 0.01 & 28.40 & 49.73 & 49.41 & 30.71 \\
    & \bf{\boa}$^{\dagger}$ & \bf{17.83} & 32.94 & 56.02 & 64.16 & 53.16 & 40.41 & 31.60 & 68.93 & 56.04 & \bf{50.41} \\
    \bottomrule
    \end{tabular}
    \begin{tablenotes}
        \item[$\dagger$] \boa \ has been applied after transforming the model via SpinQuant.
        \item[*] The LET option has been deactivated for OmniQuant and AffineQuant because this option does not support models exploiting GQA.
    \end{tablenotes}
    \end{threeparttable}
    \label{tab:weight_act_quant_llama3.2-1b}
\end{table*}

\begin{table*}[!ht]
    \renewcommand{\arraystretch}{1.0}
    \fontsize{6.25pt}{7.5pt}\selectfont
    \centering
    \caption{Weight-activation quantization performance on the transformed LLaMA3.2-3B}
    \begin{threeparttable}
    \begin{tabular}{c c c c c c c c c c c c c}
    \toprule
    \multirowcell{2}{Precision} & \multirowcell{2}{Method} & \multirowcell{2}{Wiki2 PPL ($\downarrow$)} & \multicolumn{9}{c}{Zero-shot Accuracy ($\uparrow$)} \\
    \cline{4-12}
    & & & Arc-c & Arc-e & BQ & HS & LAMB & OBQA & PIQA & WG & Average \\
    \toprule
    FP16 & Baseline & 11.04 & 46.16 & 67.80 & 78.62 & 70.44 & 62.15 & 36.00 & 75.52 & 67.40 & 63.01 \\
    \midrule
    \multirowcell{7}{W2A4KV16} 
    & OmniQuant & 5.8e3 & 25.94 & 26.81 & 37.89 & 25.83 & 0.00 & 28.80 & 50.44 & 49.80 & 30.69 \\
    & AffineQuant & 5.6e3 & 23.89 & 26.64 & 38.41 & 25.74 & 0.00 & 26.00 & 49.56 & 49.49 & 29.97 \\
    & SpinQuant-RTN & 46.31 & 24.06 & 31.14 & 51.87 & 31.41 & 12.70 & 26.60 & 55.39 & 50.99 & 35.52 \\
    & SpinQuant-GPTQ & 68.74 & 23.12 & 32.45 & 38.29 & 31.84 & 9.53 & 27.20 & 54.95 & 51.30 & 33.59 \\
    & DuQuant & 1.2e5 & 25.94 & 24.75 & 39.91 & 26.77 & 0.00 & 29.20 & 51.63 & 49.80 & 31.00 \\
    & DuQuant + LWC & 1.2e3 & 24.91 & 25.63 & 37.83 & 26.16 & 0.00 & 28.40 & 52.72 & 48.07 & 30.47 \\
    & \bf{\boa}$^{\dagger}$ & \bf{34.25} & 24.57 & 36.28 & 60.24 & 38.57 & 17.38 & 29.40 & 57.56 & 52.49 & \bf{39.56} \\
    \midrule
    \multirowcell{7}{W3A3KV16} 
    & OmniQuant & 1.2e4 & 25.00 & 26.56 & 37.95 & 25.84 & 0.00 & 27.20 & 49.73 & 51.14 & 30.43 \\
    & AffineQuant & 1.0e4 & 25.77 & 25.72 & 38.13 & 26.21 & 0.00 & 28.20 & 50.05 & 49.72 & 30.48 \\
    & SpinQuant-RTN & 36.37 & 24.23 & 35.52 & 55.66 & 36.04 & 15.19 & 26.80 & 57.78 & 49.72 & 37.62 \\
    & SpinQuant-GPTQ & 27.23 & 28.33 & 40.70 & 60.73 & 44.15 & 21.53 & 28.20 & 58.16 & 54.06 & 41.98 \\
    & DuQuant & 564.6 & 22.01 & 27.02 & 43.21 & 29.68 & 0.94 & 27.00 & 50.60 & 52.80 & 31.66 \\
    & DuQuant + LWC & 82.65 & 23.04 & 31.61 & 48.75 & 33.24 & 3.01 & 28.80 & 53.32 & 53.43 & 34.40 \\
    & \bf{\boa}$^{\dagger}$ & \bf{23.76} & 28.75 & 44.53 & 60.98 & 46.08 & 29.32 & 28.40 & 60.07 & 54.22 & \bf{44.04} \\
    \midrule
    \multirowcell{7}{W4A4KV16} 
    & OmniQuant & 131.8 & 24.57 & 34.55 & 48.75 & 36.10 & 3.21 & 27.00 & 56.37 & 51.07 & 35.20 \\
    & AffineQuant & 131.8 & 23.55 & 34.60 & 47.65 & 35.80 & 3.01 & 27.80 & 56.26 & 49.57 & 34.78 \\
    & SpinQuant-RTN & 12.42 & 38.23 & 60.94 & 72.69 & 64.68 & 55.02 & 31.60 & 71.22 & 61.09 & 56.93 \\
    & SpinQuant-GPTQ & 11.87 & 39.51 & 63.05 & 74.31 & 66.42 & 56.57 & 35.80 & 71.22 & 62.83 & 58.71 \\
    & DuQuant & 13.91 & 38.40 & 59.93 & 74.59 & 66.43 & 53.84 & 36.60 & 70.73 & 60.46 & 57.62 \\
    & DuQuant + LWC & 13.32 & 38.23 & 60.73 & 75.29 & 65.93 & 51.58 & 36.40 & 71.87 & 63.38 & 57.93 \\
    & \bf{\boa}$^{\dagger}$ & \bf{11.57} & 40.70 & 63.68 & 75.50 & 66.86 & 56.77 & 36.00 & 70.24 & 63.61 & \bf{59.17} \\
    \midrule
    \multirowcell{7}{W4A4KV4} 
    & OmniQuant & 169.1 & 23.89 & 34.22 & 43.76 & 32.36 & 1.62 & 26.20 & 54.24 & 50.51 & 33.35 \\
    & AffineQuant & 197.1 & 21.42 & 32.87 & 42.20 & 31.70 & 1.25 & 27.00 & 55.22 & 50.75 & 32.80 \\
    & SpinQuant-RTN & 12.96 & 39.16 & 61.87 & 72.42 & 63.06 & 48.84 & 34.40 & 69.70 & 58.80 & 56.03 \\
    & SpinQuant-GPTQ & 12.24 & 39.33 & 61.91 & 72.32 & 65.56 & 54.57 & 35.20 & 70.13 & 61.33 & 57.54 \\
    & DuQuant & 14.65 & 39.51 & 60.02 & 72.08 & 65.76 & 52.54 & 34.00 & 69.53 & 62.19 & 56.95 \\
    & DuQuant + LWC & 13.84 & 38.99 & 59.68 & 72.05 & 64.76 & 49.18 & 35.60 & 70.40 & 61.56 & 56.53 \\
    & \bf{\boa}$^{\dagger}$ & \bf{11.98} & 39.08 & 63.64 & 74.22 & 65.60 & 55.51 & 38.80 & 71.22 & 63.14 & \bf{58.90} \\
    \bottomrule
    \end{tabular}
    \begin{tablenotes}
        \item[$\dagger$] \boa \ has been applied after transforming the model via SpinQuant.
        \item[*] The LET option has been deactivated for OmniQuant and AffineQuant because this option does not support models exploiting GQA.
    \end{tablenotes}
    \end{threeparttable}
    \label{tab:weight_act_quant_llama3.2-3b}
\end{table*}

\newpage

\section{Pseudocode for GPTQ}  \label{appendix:pseudocode-gptq}

In this appendix, we provide the pseudocode of the conventional GPTQ~\citep{frantar2023optq}, which is omitted in the main manuscript due to the page limitation.

\begin{algorithm*}[!htb]
\begin{spacing}{1.1}
\caption{GPTQ} 
\label{algo:gptq}
\renewcommand\algorithmicrequire{\textbf{Input}:}
\renewcommand\algorithmicensure{\textbf{Output}:}
\begin{algorithmic}[1]
\REQUIRE weights $\mathbf{W}$, Hessian information $\mathbf{U}_{\text{col}}$, and pre-determined step size $\mathbf{S}$
    \STATE Initialize quantized output: $\mathbf{Q} \leftarrow \mathbf{0}_{d_{\text{row}} \times d_{\text{col}}}$
    \STATE Initialize quantization errors: $\mathbf{E} \leftarrow \mathbf{0}_{d_{\text{row}} \times d_{\text{col}}}$
    \FOR{$j=0, \cdots, d_{\text{col}} - 1$}
    \STATE Quantize the $j$-th column: $\mathbf{Q}_{:, j} \leftarrow \text{quant} ( \mathbf{W}_{:, j}, \mathbf{S})$
    \STATE Estimate quantization error: $\mathbf{E}_{:, j} \leftarrow (\mathbf{W}_{:, j} - \mathbf{Q}_{:, j}) / [\mathbf{U}_{\text{col}}]_{j, j}$
    \STATE Update weights: $\mathbf{W}_{:, j:} \leftarrow \mathbf{W}_{:, j:} - \mathbf{E}_{:, j} \cdot [\mathbf{U}_{\text{col}}]_{j, j:}$ 
    \ENDFOR
\ENSURE quantized weights $\mathbf{Q}$, quantization error $\mathbf{E}$
\end{algorithmic}
\end{spacing}
\end{algorithm*}

\end{document}